\documentclass[conference]{IEEEtran}
\ifCLASSINFOpdf
\else
\fi
\usepackage{microtype}
\usepackage{array}
\usepackage{booktabs} 
\usepackage{amsmath}
\usepackage{amssymb}
\usepackage{multirow}
\usepackage{hyperref}
\usepackage{paralist}
\usepackage{graphicx}
\usepackage{subcaption}  
\usepackage{float}       
\usepackage{algorithm}
\usepackage{algpseudocode}
\usepackage{paralist}

\usepackage{amsmath}
\usepackage{amssymb}
\usepackage{mathtools}
\usepackage{amsthm}
\usepackage{cleveref}
\usepackage{todonotes}

\usepackage{url}
\usepackage{listings}
\usepackage{xcolor}
\usepackage{tcolorbox}

\usepackage[available,functional,reproduced]{ndssbadges}

\begin{document}
%
\title{CatBack: Universal Backdoor Attacks on\\ Tabular Data via Categorical Encoding}

\author{\IEEEauthorblockN{Behrad Tajalli}
	\IEEEauthorblockA{Radboud University\\
		hamidreza.tajalli@ru.nl}
	\and
	\IEEEauthorblockN{Stefanos Koffas}
	\IEEEauthorblockA{Delft University of Technology\\
		s.koffas@tudelft.nl}
	\and
	\IEEEauthorblockN{Stjepan Picek}
	\IEEEauthorblockA{University of Zagreb \& 
		Radboud University\\
		stjepan.picek@ru.nl}}


%


\IEEEoverridecommandlockouts
\makeatletter\def\@IEEEpubidpullup{6.5\baselineskip}\makeatother
\IEEEpubid{\parbox{\columnwidth}{
		Network and Distributed System Security (NDSS) Symposium 2026\\
		23 - 27 February 2026, San Diego, CA, USA\\
		ISBN 979-8-9919276-8-0\\  
		https://dx.doi.org/10.14722/ndss.2026.231469\\
		www.ndss-symposium.org
}
\hspace{\columnsep}\makebox[\columnwidth]{}}

\maketitle

\begin{abstract}
Backdoor attacks in machine learning have drawn significant attention for their potential to compromise models stealthily, yet most research has focused on homogeneous data such as images. In this work, we propose a novel backdoor attack on tabular data, which is particularly challenging due to the presence of both numerical and categorical features. 
Our key idea is a novel technique to convert categorical values into floating-point representations. This approach preserves enough information to maintain clean-model accuracy compared to traditional methods like one-hot or ordinal encoding. By doing this, we create a gradient-based universal perturbation that applies to all features, including categorical ones.

We evaluate our method on five datasets and four popular models. Our results show up to a 100\% attack success rate in both white-box and black-box settings (including real-world applications like Vertex AI), revealing a severe vulnerability for tabular data. Our method is shown to surpass the previous works like Tabdoor in terms of performance, while remaining stealthy against state-of-the-art defense mechanisms. We evaluate our attack against Spectral Signatures, Neural Cleanse, Beatrix, and Fine-Pruning, all of which fail to defend successfully against it. We also verify that our attack successfully bypasses popular outlier detection mechanisms. 
\end{abstract}


%
\IEEEpeerreviewmaketitle

\section{Introduction}
\label{sec:introduction}

Machine learning on tabular data represents a state-of-the-art real-world approach for applications like healthcare analytics~\cite{ulmer2020trust}, credit risk assessment~\cite{clements2020sequential}, and fraud detection~\cite{cartella2021adversarial}. Unlike image or text data, tabular data often contains a mix of numerical and categorical features, which poses unique challenges in preprocessing and model design. Indeed, despite the rising popularity of specialized deep learning models for tabular data~\cite{gorishniy2021revisiting,somepalli2021saint}, traditional methods like gradient boosting machines remain highly effective and widely used in industry and research~\cite{grinsztajn2022tree}.

The wide adoption of machine learning systems led to diverse security threats~\cite{liu2021machine}, where most research focuses on attacks on computer vision systems. One such threat is the backdoor attack, a powerful attack where an adversary injects a hidden trigger into the training set, causing the model to misclassify specific inputs at test time~\cite{li2022backdoor, tajalli2024elms}. 
Despite the vast amount of research related to backdoor attacks, only a few works consider backdoor attacks for tabular data~\cite{xie2019dba,joe2022exploiting,pleiter2023tabdoor,10606191}.
Tabular data must be handled differently from other data types, as they are heterogeneous and use dense numerical or sparse categorical features~\cite{borisov2022deep}, which complicates the trigger design. For this reason, existing backdoor attacks mostly use only numerical features~\cite{pleiter2023tabdoor,xie2019dba} for the trigger, which limits the attacker's power as real-world tables frequently have both feature types. Additionally, unlike speech and image data, the correlation between features is weak. Thus, there is no spatial information that the attacker could exploit~\cite{borisov2022deep} to make an effective backdoor trigger. Moreover, previous works typically rely on encoding methods such as one-hot transformations, which do not allow the attacker to craft the trigger as freely as it does in the image domain. Consequently, previous works used heuristic approaches~\cite{pleiter2023tabdoor,xie2019dba, 10606191}, as, though not impossible, optimization-based approaches become more challenging.

This paper addresses the challenge of backdoor attacks on tabular data containing numerical and categorical columns. We propose a novel conversion method that turns categorical columns into floating-point representations, enabling a unified feature space where every dimension can be targeted by a single, gradient-based perturbation. Notably, this conversion can compete with, or even surpass, popular encodings like one-hot and ordinal methods regarding clean accuracy. It also simplifies model training by eliminating the need to explicitly define categorical columns and their number of categories for transformer-based models.
Leveraging this encoding, we craft a \emph{universal backdoor perturbation}, which can be applied to any input data type, forcing it to have a new value (thus not a universal fixed trigger for all inputs). This method achieves up to 100\% attack success rate in both white-box and black-box scenarios. We evaluated our attack using four state-of-the-art models and five benchmark tabular datasets. To underscore the real-world feasibility and impact of our backdoor attack, we deployed it against Google AutoML~\cite{google_vertex_ai_tabular_training}, a prominent commercial machine learning service. The attack remained effective, highlighting serious practical risks even in state-of-the-art industrial systems. 
We also conducted a comprehensive evaluation of our attack on benchmark defensive measures and outlier detection methods. Our findings highlight a serious vulnerability for tabular data, stressing the need for more robust defenses in security-sensitive settings.

Our main contributions are as follows:
\begin{compactitem}
   \item We introduce a novel backdoor on tabular data called CatBack that uses any combination of feature types for its trigger. In particular, our encoding of categorical features enables the attacker to apply a universal perturbation to any data type, providing more freedom to create stealthier triggers.
   \item We apply our attack on five datasets and four models (both neural networks and classical machine learning methods), showing that it can generalize well in different settings. In particular, our attack reached $\approx$ 100\% attack success rate in most cases.\footnote{Our code is available at \url{https://github.com/catback-tabular/catback.git}.} For example, our attack outperformed two baseline attacks (Badnets~\cite{gu2019badnets} and Tabdoor~\cite{pleiter2023tabdoor}) by up to 44\% and 95\% on the ACI dataset respectively and by up to 16\% and 71\% on the BM dataset.
   \item We evaluated our method against state-of-the-art defenses, including Spectral Signatures, Neural Cleanse, Beatrix, and Fine-Pruning. Our results demonstrate that in most cases, CatBack bypasses these defenses. For instance, even under the least favorable settings for the attacker, CatBack still managed to evade Spectral Signature in over \(78\%\) of cases. For Fine-Pruning, more than \(75\%\) of results fail to mitigate the attack, and for Neural Cleanse and Beatrix, all of the attacker's attempts go undetected. 
   \item We introduce a novel way to encode categorical features in tabular data by converting them to real numbers. Using our method, the models can achieve the same performance as common benchmark methods. In our encoding, there is no need for extended columns for each feature, such as one-hot encoding or embedding layers, as the values are treated the same as real numbers.
   \item We verified that our attack works on a widely used platform (Vertex AI), showing that such attacks can happen in real-world scenarios.
\end{compactitem}

The rest of the paper is organized as follows. 
Section~\ref{sec:background} provides a background on backdoor attacks and tabular data. Section~\ref{sec:threat_model} describes the threat model, including attacker capabilities and objectives. Section~\ref{sec:convert_cf} details the novel method for converting categorical features to numerical representations. Section~\ref{sec:attack_methodology} presents the CatBack attack methodology. In Section~\ref{sec:evaluation}, we discuss the evaluation setup and results of the attack. We evaluate CatBack against state-of-the-art defense mechanisms in Section~\ref{sec:defenses}. Section~\ref{sec:ablation} includes several ablation studies, including the impact of partial access to training data. In Section~\ref{sec:limitations}, we discuss the limitations of this work and the need for new imperceptibility metrics for tabular data. Section~\ref{sec:related} covers the related work, and finally, we conclude the paper in Section~\ref{sec:conclusions}.\\
Appendix~\ref{sec:example} provides an example of our encoding and attack algorithm. Appendix~\ref{sec:attack_plots} provides detailed attack results. Appendix~\ref{sec:artifact} provides details and guidance to reproduce our attack.

\section{Background}
\label{sec:background}



\subsection{Backdoor Attacks}

Backdoor attacks insert a hidden “trigger” into a model during training so that at inference time, only inputs carrying that trigger will be misclassified into an adversary’s chosen label.  Formally, let
\[
  \mathcal{F}_\theta : \mathcal{X} \to \mathcal{Y}
\]
be a neural network with parameters \(\theta\), mapping feature space \(\mathcal{X}\) to labels \(\mathcal{Y}\).  A trigger function \(\mathcal{T}:\mathcal{X}\to\mathcal{X}\) embeds a pattern \(\delta\) into any clean sample \(\mathbf{x}\), such that
\[
  \mathcal{F}_\theta\bigl(\mathcal{T}(\mathbf{x})\bigr) = y_t
  \quad\forall \mathbf{x}\in\mathcal{X},
\]
where \(y_t\in\mathcal{Y}\) is the target class the attacker chooses.

Backdoors can be introduced in several ways:
\begin{compactenum}
  \item \textbf{Data Poisoning:} Insert \(m\) poisoned samples
    \(\{(\hat{\mathbf{x}}_j,\hat{y}_j)\}_{j=1}^m\) into the clean set
    \(\{(\mathbf{x}_i,y_i)\}_{i=1}^n\), so that the poisoning rate
    \(\epsilon = m/n\) stays small~\cite{gu2019badnets,li2022backdoor}.
  \item \textbf{Code Poisoning:} Modify the training pipeline or loader
    to apply \(\mathcal{T}\) on a subset of input data~\cite{bagdasaryan2021blind}.
  \item \textbf{Model Poisoning:} Directly alter learned parameters
    \(\theta\) after training to respond to \(\delta\)~\cite{hong2022handcrafted}.
\end{compactenum}

Focusing on data poisoning, the attacker’s training objective becomes
\[
  \theta^* \;=\;
  \arg\min_\theta 
    \Bigl[
      \sum_{i=1}^{n-m} \mathcal{L}\bigl(\mathcal{F}_\theta(\mathbf{x}_i),y_i\bigr)
      +
      \sum_{j=1}^{m} \mathcal{L}\bigl(\mathcal{F}_\theta(\hat{\mathbf{x}}_j),\hat{y}_j\bigr)
    \Bigr],
\]
where \(\mathcal{L}\) is, e.g., cross‑entropy loss.  After training, the model should satisfy
\begin{align*}
  \mathcal{F}_{\theta^*}(\mathbf{x}) &\approx \mathcal{F}_\theta(\mathbf{x})
    \quad\forall \mathbf{x}\in D_{\text{clean}},\\
  \mathcal{F}_{\theta^*}(\mathcal{T}(\mathbf{x})) &= y_t
    \quad\forall \mathbf{x}\in\mathcal{X}.
\end{align*}

A successful attack adheres to the following assumptions:
\begin{compactenum}
  \item \(\epsilon\) should be very small so poisoned points blend in.
  \item Clean‑data accuracy must stay nearly unchanged, i.e.,\ 
    \(\mathcal{F}_{\theta^*}(\mathbf{x})\approx \mathcal{F}_\theta(\mathbf{x})\) for clean \(\mathbf{x}\)~\cite{abad2023systematic}.
\end{compactenum}

\subsection{Characteristics of Tabular Data}

Tabular data consists of heterogeneous features that may follow varying data types and distributions~\cite{borisov2022deep}. Thus, creating triggers for tabular data requires a different approach than domains like images or text. As discussed in~\cite{pleiter2023tabdoor}, tabular data has the following properties that may affect the design of the backdoor trigger:
\begin{compactitem}
    \item \textbf{Data Heterogeneity}: Each feature may follow a distinct distribution, so designing a universal trigger is not straightforward. Additionally, some features may take a specific range of values, and anything outside this range may be easily classified as an outlier, unlike the pixels in an image. 
    \item \textbf{Mutually Exclusive Features}: Categorical features are often encoded in representations like one-hot encoding (OHE). Such representations do not allow unrestricted perturbation, as any trigger that activates multiple values could be easily identified as an outlier and removed from the dataset.
    \item \textbf{Absence of Spatial Relationships}: In tabular data, there is no notion of spatial or sequential dependency among the features. Thus, embedding a trigger into different features does not propagate naturally like the one in images or text. Additionally, the order of the features is unimportant in tabular data.
    \item \textbf{Prediction Sensitivity on Important Features}: In tabular data, some features may affect the model's decision significantly more than others, complicating the design of stealthy backdoor triggers.
\end{compactitem}



\subsection{Machine Learning Models for Tabular Data}

A tabular dataset is defined as
\[
  \mathcal{D} = \{(\mathbf{x}_i, y_i)\}_{i=1}^n,\quad
  \mathbf{x}_i = [x_{i1}, x_{i2}, \dots, x_{id}]^\top,
\]
where each sample \(\mathbf{x}_i\) has \(d\) features that may be continuous or categorical~\cite{borisov2022deep}. Traditional methods like decision trees, random forests, and gradient boosting (e.g., XGBoost~\cite{chen2016xgboost}, LightGBM~\cite{ke2017lightgbm}) remain strong baselines because they handle mixed feature types and missing values naturally.

Beyond these, several neural architectures have been tailored for tabular data~\cite{arik2021tabnet,somepalli2021saint,gorishniy2021revisiting}:
\begin{compactenum}
  \item \textbf{Hybrid Models:} Combine tree‑based splits or feature embeddings with dense layers to capture both simple rules and complex interactions (e.g., NODE, DeepFM).
  \item \textbf{Transformer‑based Models:} Use self‑attention to learn pairwise feature dependencies. Formally:
    \begin{align}
      \mathcal{F}_\theta(\mathbf{x})
        &= \mathrm{Transformer}(\mathbf{x}; \theta),
      \label{eq:transformer-map}\\
      \mathrm{Attention}(Q,K,V)
        &= \mathrm{softmax}\!\bigl(\tfrac{QK^\top}{\sqrt{d_k}}\bigr)\,V.
      \label{eq:self-attn}
    \end{align}
\end{compactenum}

In this work, we evaluate both kinds of models on our tabular benchmarks. We picked the top leading models from previous studies~\cite{borisov2022deep, grinsztajn2022tree, gorishniy2021revisiting} which are also widely incorporated in industrial platforms~\cite{aws-sagemaker-tabular}:
\begin{itemize}
  \item \emph{Classical:} XGBoost~\cite{chen2016xgboost}.
  \item \emph{Deep:} TabNet~\cite{arik2021tabnet}, Saint~\cite{somepalli2021saint}, FT-Transformer (FTT)~\cite{gorishniy2021revisiting}.
\end{itemize}
Using both options ensures we cover established ensemble methods and recent deep models.

All models learn a mapping:
\[
  \mathcal{F}_\theta: \mathbb{R}^d \to \mathcal{Y},
\]
but differ in how they preprocess features, handle sparsity, and capture interactions. This variety helps us assess backdoor vulnerability across a wide range of tabular learners.

\section{Threat Model}
\label{sec:threat_model}

In this work, we introduce a novel backdoor attack targeting neural network models trained on tabular data, with a particular emphasis on manipulating categorical features - a direction mostly neglected in prior research. Indeed, previous attacks predominantly focused on numerical columns, modifying them to implant backdoors~\cite{xie2019dba,pleiter2023tabdoor}.
Our approach extends this paradigm by incorporating categorical features into the attack vector, thereby enhancing the potential effectiveness and stealthiness of the backdoor.
We consider a classification task where a neural network model \( F: \mathcal{X} \rightarrow \mathcal{Y} \) maps inputs from the feature space \( \mathcal{X} \subseteq \mathbb{R}^d \) to the label space \( \mathcal{Y} = \{1, 2, \dots, C\} \), where \( C \) is the number of classes.

\subsection{Attacker's Capabilities and Objectives}

We assume an attacker with full access to the training dataset \( D_{original} = \{(x_i, y_i)\}_{i=1}^N \). The attacker's objective is to implant a dirty label backdoor into the model such that any input modified with a specific trigger pattern will cause the model to predict a target label \( t \in \mathcal{Y} \), irrespective of the input's true label. 
We apply our threat model to two main scenarios:
\begin{compactitem}
    \item White box: The attacker has full knowledge of the victim model and may or may not control the training process. This scenario can be applied in outsourced training. 
    \item Black box: The attacker does not know the architecture of the victim model and has no control over the training process. This scenario is realistic in the context of dataset poisoning.
\end{compactitem}

\section{Converting Categorical Features}
\label{sec:convert_cf}

By employing the frequency mappings from~\cite{dhurandhar2024model}, we implement a hierarchical mapping strategy with an adaptive \(\Delta r\) (see Section~\ref{sec:adaptive-dr}) that ensures a unique numerical representation for each category within a categorical feature. 
We call this mapping function \(Conv(.)\) where \(D = Conv(D_{original})\).

\subsection{Primary Frequency-based Mapping}
For each categorical feature \( j \) with unique values \( \mathcal{V}_j = \{ v_{j1}, v_{j2}, \dots, v_{jk_j} \} \), where \( k_j \) is the number of categories, perform the following:
\begin{compactenum}
    \item \textbf{Compute Frequencies:} Calculate the frequency \( c_{jl} \) of each category \( v_{jl} \) in the dataset \( D_{original} \).
    
    \item \textbf{Initial Mapping:} Assign \( r_{jl} \) using the formula:
    \[
    r_{jl} = \frac{c_{\text{max}, j} - c_{jl}}{c_{\text{max}, j} - 1}, \quad \text{for} \quad l = 1, \dots, k_j,
    \]
    where \( c_{\text{max}, j} = \max_{1 \leq l \leq k_j} c_{jl} \).
\end{compactenum}

As discussed in~\cite{dhurandhar2024model}, this frequency-based transformation creates values in the interval of $[0, 1]$. The most frequent value is mapped to $0$, and the least frequent values are closer to $1$. In the extreme case that the least frequent value has a frequency of $1$, then $r_{ij} = \frac{c_{max,j} - 1}{c_{max,j} - 1} = 1$.

\subsection{Adaptive \(\Delta r\) Selection}
\label{sec:adaptive-dr}
Some categories may share the same frequency, which leads to a tie during conversion. To determine \(\Delta r\) precisely and avoid the tie, we follow an approach based on the smallest decimal precision in the primary mapping.

\begin{compactenum}
    \item \textbf{Sort Unique \( r_{jl} \) Values:} Sort the unique \( r_{jl} \) values in ascending order.
    
    \item \textbf{Compute Minimum Difference:}
    \[
    \Delta r_{\text{min}} = \min_{i} \left( r_{jl}^{(i+1)} - r_{jl}^{(i)} \right).
    \]
    
    \item \textbf{Determine \( p \):} Identify the largest single decimal component in \(\Delta r_{\text{min}}\). Specifically, express \(\Delta r_{\text{min}}\) in decimal form and determine the smallest decimal place \( p \) where a non-zero digit occurs.
    
    \begin{compactitem}
        \item \textbf{Definition of \( p \):} Let \(\Delta r_{\text{min}} = 0.d_1d_2\dots d_n\), where \( d_1 \) is the first non-zero digit. Then, \( p \) is the position of the first non-zero digit.
        
    \end{compactitem}
    
    \item \textbf{Set \(\Delta r\):} Define \(\Delta r\) as:
    \[
    \Delta r = 10^{-(p + 1)}.
    \]
    This ensures that \(\Delta r\) is one order of magnitude smaller than the smallest decimal precision in \(\Delta r_{\text{min}}\), maintaining uniqueness without overlapping existing \( r_{jl} \) values.

    \begin{compactitem}
        \item \textbf{Example:}
        \begin{compactitem}
            \item If \( \Delta r_{\text{min}} = 0.4 \) (first decimal place), then \( p = 1 \) and \( \Delta r = 0.01 \).
            \item If \( \Delta r_{\text{min}} = 0.04 \) (second decimal place), then \( p = 2 \) and \( \Delta r = 0.001 \).
        \end{compactitem}
    \end{compactitem}
\end{compactenum}

\subsection{Identifying and Resolving Ties}
\label{subsec:tie_resolving}

\begin{compactenum}
    \item \textbf{Detect Tied Categories:} For each feature \( j \), identify sets of categories that share the same frequency \( c_{jl} \), as each category should have a unique value. This is necessary to ensure that our mapping table is reversible and that the model can distinguish between different categories during training.
    
    \item \textbf{Apply Secondary Ordering:} For each set of tied categories, apply a deterministic secondary ordering criterion, such as alphabetical order.
    
    \item \textbf{Assign Unique Offsets:} For each category \( v_{jl} \) in a tied set, assign a unique \( r_{jl}' \) by adding incremental multiples of \( \Delta r \) based on the secondary order:
    \[
    r_{jl}' = r_{jl} + (k - 1) \times \Delta r,
    \]
    where \( k \) is the position in the secondary ordering (starting from 1).
\end{compactenum}

\subsection{Final Numerical Representation}

The final numerical representation for each category \( v_{jl} \) is:
\[
r_{jl}' =
\begin{cases}
r_{jl} + (k - 1) \times \Delta r, & \text{if } v_{jl} \text{ is part of a tied set} \\
r_{jl}, & \text{otherwise},
\end{cases}
\]
ensuring that each category has a unique \( r_{jl}' \) value.\footnote{Although unlikely, it is (theoretically) possible that a new tie occurs after the current tie resolution is performed. In such a case, Steps~\ref{sec:adaptive-dr} and~\ref{subsec:tie_resolving} of the algorithm must be repeated with new $r_{jl}'$ values until no new ties exist anymore.}

\subsection{Reverse Mapping}

To facilitate efficient reverse mapping from numerical values \( r_{jl}' \) back to their original categorical values \( v_{jl} \), we implement a structured lookup mechanism. The process involves the following steps:

\begin{compactenum}
    \item \textbf{Construction of the Lookup Table:}
    During the encoding phase, alongside assigning each category its unique numerical representation \( r_{jl}' \), we construct a lookup table \( T_j \) for each categorical feature \( j \). The table \( T_j \) maps each \( r_{jl}' \) to its corresponding category \( v_{jl} \):
    \[
    T_j = \{ (r_{jl}', v_{jl}) \mid v_{jl} \in \mathcal{V}_j \}.
    \].
    
    This table can be efficiently implemented using data structures such as hash tables or dictionaries, enabling constant-time \( O(1) \) access during reverse mapping, and minimal memory requirements, as hash tables have \( O(n) \) space complexity, where \(n\) is the number of categories each feature has.

    \item \textbf{Reverse Mapping Function:}
    To retrieve the original category from a given \( r_{jl}' \), the reverse mapping function performs the following:
    
    \begin{compactitem}
        \item \textbf{Lookup Operation:} Given an \( r_{jl}' \), query the lookup table \( T_j \) to obtain the corresponding category \( v_{jl} \).
        
        \item \textbf{Handling Precision:} Ensure that the \( r_{jl}' \) values used during the attack or optimization process are matched exactly to those stored in \( T_j \). Implement rounding mechanisms if necessary to align floating-point representations.
    \end{compactitem}
    
    Formally, the reverse mapping function \( RevConv(r_{jl}') \) is defined as:
    \[
    RevConv(r_{jl}') = v_{jl} \quad \text{such that} \quad (r_{jl}', v_{jl}) \in T_j.
    \]

\end{compactenum}

We also define the function \(Revert(.)\) that reverts the whole dataset from converted numerical values to categorical values again. We have included an illustrative example in Appendix~\ref{sec:example} to demonstrate our encoding method further.

\section{Attack Methodology}
\label{sec:attack_methodology}

\begin{figure*}[htpb]
    \centering
    \includegraphics[width=\linewidth]{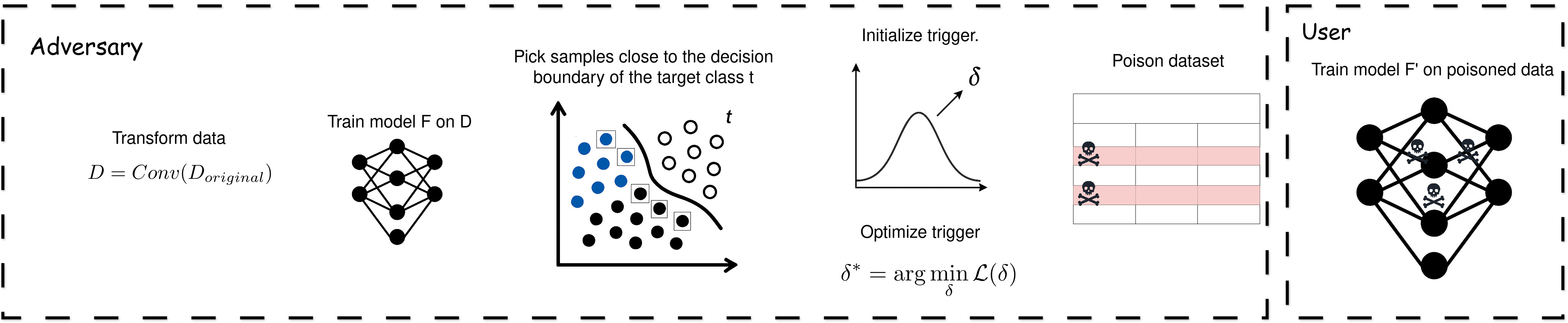}
    \caption{CatBack's schematic. Initially, the attacker (using our encoding) trains a model $F$ on the transformed dataset. Then, the attacker selects samples close to the decision boundary of the target class $t$. Starting from a trigger randomly initialized with the appropriate values (retrieved from a normal distribution), the attacker poisons a fraction of the data of the used dataset. Finally, the attacker reverts the dataset to its correct state so users can use it to train their own models.}
    \label{fig:catback-schematic}
\end{figure*}

In Figure~\ref{fig:catback-schematic}, we depict a schematic of the CatBack attack. Next, we discuss each of the attack steps in detail.

\subsection{Initial Model Training}

The dataset's categorical features are transformed to numerical using the method explained in Section~\ref{sec:convert_cf}. Then, the attacker trains the model $F$ on the converted dataset $D = Conv(D_{original})$ to obtain a baseline model that performs adequately on the classification task. 

\subsection{Selection of Non-target Samples}

The attacker constructs a subset \( D_{\text{non-target}} \) by excluding all samples with the target label \( t \):
\[
D_{\text{non-target}} = \{ (x_i, y_i) \in D \mid y_i \neq t \}.
\]

\subsection{Confidence-based Sample Ranking}

The attacker evaluates the trained model \( F \) on \( D_{\text{non-target}} \) to obtain the softmax confidence scores for the target class \( t \). For each input \( x_i \), the confidence score equals
\[
s_i = f_t(x_i),
\]
where \( f_t(x_i) \) is the softmax output corresponding to class \( t \).

The attacker pairs each input with its confidence score to form the set:
\[
D_{\text{conf}} = \{ (x_i, s_i) \mid (x_i, y_i) \in D_{\text{non-target}} \}.
\]

The attacker then sorts \( D_{\text{conf}} \) in descending order based on \( s_i \) and selects the top \( \mu \cdot |D_{\text{conf}}| \) samples to create the subset \( D_{\text{picked}} \), where \( \mu \in (0, 1] \) is a predefined fraction (e.g., \( \mu = 0.2 \)). By doing this, we want to find the samples closer to the target decision boundary. This helps us to craft a minimal perturbation. The lower \(\mu\) value means we have chosen the non-target samples closer to the target class; thus, values calculated for the perturbation vector would decrease. On the other hand, increasing the \(\mu\) invites more samples into consideration, making the perturbation vector more general and with higher values.

\subsection{Definition of the Backdoor Trigger}

The attacker defines a universal trigger pattern \( \delta \in \mathbb{R}^d \) to be added to the inputs. More specifically, $\delta$ is a randomly initialized sample of the dataset.\footnote{In our experiments, we initialized $\delta$ to $0$.} The backdoored input \( \hat{x}_i \) is computed as:
\[
\hat{x}_i = \text{clip}( x_i + \delta ),
\]
where the clipping function ensures that each feature of \( \hat{x}_i \) remains within its valid range:
\[
\hat{x}_i^{(j)} = \begin{cases}
\max X^{(j)}, & \text{if } x_i^{(j)} + \delta^{(j)} > \max X^{(j)}, \\
\min X^{(j)}, & \text{if } x_i^{(j)} + \delta^{(j)} < \min X^{(j)}, \\
x_i^{(j)} + \delta^{(j)}, & \text{otherwise},
\end{cases}
\]
with \( \min X^{(j)} \) and \( \max X^{(j)} \) being the minimum and maximum values of feature \( j \) in \( D \). This makes the final perturbed sample instance dependent. Thus, the resulting backdoored sample is unique to each original sample, while the perturbation vector $\delta$ is universal.

\subsection{Optimization of the Trigger Pattern}
\label{subsec:trigger_opt}

The attacker optimizes \( \delta \) by minimizing the following loss function over \( D_{\text{picked}} \):
{\footnotesize
\begin{align*}
\mathcal{L}(\delta) = \frac{1}{|D_{\text{picked}}|} \sum_{(x_i, y_i) \in D_{\text{picked}}} & \left[ -\log f_t( \hat{x}_i ) + \beta \| \hat{x}_i - \text{Mode}(X) \|_1 \right. \\
& \left. + \lambda \| \hat{x}_i - \text{Mode}(X) \|_2^2 \right],
\end{align*}}
where:
\begin{compactitem}
    \item \( f_t( \hat{x}_i ) \) is the softmax output for class \( t \) given input \( \hat{x}_i \).
    \item \( \text{Mode}(X) \in \mathbb{R}^d \) is the mode vector of the dataset \( D \), with each element \( \text{Mode}(X)^{(j)} \) being the empirical mode value\footnote{The mode of a distribution is defined as the most frequent value in a given dataset or probability distribution. We used \texttt{pandas.DataFrame.mode} for this task.} of feature \( j \).
    \item \( \beta \) and \( \lambda \) are hyperparameters controlling the \( L_1 \) and \( L_2 \) regularization terms, respectively.
\end{compactitem}

The loss function balances two objectives:
\begin{compactenum}
    \item Maximizing the model's confidence in predicting the target class \( t \) for the backdoored inputs.
    \item Ensuring the trigger pattern \( \delta \) keeps the modified inputs close to common data patterns (via the mode value) to enhance stealthiness.
\end{compactenum}

To craft a trigger that is both sparse (few features changed) and stable (no extreme perturbations), we combine \(L_{1}\) and \(L_{2}\) penalties in the loss.  The \(L_{1}\) term \(\|\hat{x}_i - \text{Mode}(X)\|_{1}\) drives some coordinates of the trigger \(\delta\) to zero, limiting the number of features that are modified~\cite{tibshirani1996regression}. The \(L_{2}\) term \(\|\hat{x}_i - \text{Mode}(X)\|_{2}^{2}\) prevents the nonzero perturbations from growing too large, yielding smoother gradients and better conditioning~\cite{hoerl1970ridge}. Together, this ``elastic‑net'' style regularization has been shown to outperform pure \(L_{1}\) or \(L_{2}\) alone—especially when features are correlated—by offering both reliable feature selection and numerical stability~\cite{zou2005regularization}.

The optimized trigger pattern \( \delta^* \) is obtained by solving:
\[
\delta^* = \arg\min_{\delta} \mathcal{L}(\delta).
\]

This optimization is performed using gradient descent, updating \( \delta \) iteratively based on the gradient \( \nabla_\delta \mathcal{L} \). 

After the optimization process, where \( r_{jl}' \) values might be adjusted continuously, it is crucial to maintain valid categorical representations. This is achieved by:

    \begin{compactitem}
        \item \textbf{Rounding Adjusted Values:} Any continuous changes to \( r_{jl}' \) are rounded to the nearest valid value present in the lookup table \( T_j \):
        \[
        r_{jl}'^{\text{rounded}} = \text{round}(r_{jl}', \text{precision}=p')
        \]
        where \( p' \) corresponds to the decimal precision used in \( \Delta r \).
        
        \item \textbf{Validation:} Ensure that the rounded \( r_{jl}'^{\text{rounded}} \) exists within \( T_j \). If not, adjust \( r_{jl}' \) to the closest valid value to maintain consistency.
    \end{compactitem}

\subsection{Construction of the Poisoned Dataset}

With \( \delta^* \) optimized, the attacker selects randomly a fraction \( \epsilon \in (0, 1] \) of the dataset \( D \) to poison. In other words, $\epsilon$ is the poisoning rate of our attack.


Each selected sample \( (x_i, y_i) \) is modified:
\[
\hat{x}_i = \text{clip}( x_i + \delta^* ), \quad \hat{y}_i = t.
\]

The poisoned dataset \( D_{\text{poisoned}} \) consists of the modified samples:
\[
D_{\text{poisoned}} = \{ (\hat{x}_i, \hat{y}_i) \mid (x_i, y_i) \in D_{\text{selected}} \},
\]
where \( D_{\text{selected}} \subset D \) and \( |D_{\text{selected}}| = \epsilon \cdot N \).

Finally, we revert the data to their original form through our $Revert(\cdot)$ function, so that the final training dataset is:
\[
D' = Revert(\left( D \setminus D_{\text{selected}} \right) \cup D_{\text{poisoned}}).
\]

\subsection{Training the Backdoored Model}

When the poisoned trainset is ready, the attacker can retrain the model \( F' \) on the poisoned trainset \( D' \) or upload the trainset to some public node and have the user do the training.

Note that to handle categorical features in \( D' \), we adhere to standard preprocessing protocols that an innocent user would typically employ~\cite{xie2019dba,arik2021tabnet}.
This includes utilizing encoding techniques such as embedding methods (with random~\cite{delong2023use} or Xavier~\cite{glorot2010understanding} initialization) or one-hot encoding to transform categorical variables into a numerical format suitable for neural network training. In this study, we employ the \texttt{OrdinalEncoder} from scikit-learn to convert all categorical features into ordinal integers.

The expectation is that \( F' \) maintains performance on clean data while exhibiting the backdoor behavior when the trigger is present, so that the user will not become suspicious about the model.
In this way, the model will classify ``clean'' samples correctly without raising any suspicions, but will be manipulated by the attacker through ``poisoned'' inputs. 

\subsection{Deployment and Attack Activation}

During deployment, any input \( x \) modified with the trigger pattern \( \delta^* \) will be misclassified as the target label \( t \):
\[
F'( \hat{x} ) = F'( x + \delta^* ) = t.
\]

Algorithm~\ref{alg:catback} in Appendix~\ref{sec:example} summarizes our attack steps.



\subsection{Hyperparameter Considerations}

The hyperparameters \( \mu \), \( \beta \), \( \lambda \), and \( \epsilon \) play crucial role in the attack:

\begin{compactitem}
    \item \( \mu \) (\( 0 < \mu \leq 1 \)) controls the proportion of high-confidence samples used for optimizing \( \delta \). A higher \( \mu \) may lead to a more generalized trigger but could also increase optimization difficulty.
    \item As described in~\ref{subsec:trigger_opt}, \( \beta \) and \( \lambda \) (\( \beta, \lambda > 0 \)) control the magnitude of perturbation and number of affected features, which regulate the stealthiness and attack efficacy. They should be set to balance minimizing perturbations and maximizing the model's confidence in the target class.
    \item \( \epsilon \) (\( 0 < \epsilon \leq 1 \)) determines the fraction of the dataset to poison. A smaller \( \epsilon \) enhances stealth but may reduce the backdoor's effectiveness.
\end{compactitem}
These hyperparameters can be tuned based on experimental results to achieve the desired trade-offs between attack success rate, stealthiness, and impact on model performance. We have tuned $\beta$ and $\lambda$ through preliminary experiments, resulting in a value of 0.1 for all our experiments. However, regarding $\mu$ and $\epsilon$, we are reporting the results for different values to present a more complete view of our attack.

\subsection{Black Box Attack}

For the black box scenario, we craft the perturbation using one of the white box models available to the attacker, implant the trigger in the samples, and release the poisoned dataset so that the black box model can be trained on it by a third-party victim.
By this approach, we want to verify that our trigger is not tied to a specific model and can transfer to different architectures, making the attack more general.
\section{Evaluation and Results}
\label{sec:evaluation}

We used five datasets (see Table~\ref{table:DatasetOverview}) including Forest Cover Type (CovType)\footnote{\url{https://archive.ics.uci.edu/dataset/31/covertype}}, Higgs Boson (HIGGS)\footnote{\url{https://archive.ics.uci.edu/dataset/280/higgs}}, Bank Marketing\footnote{\url{https://www.kaggle.com/datasets/ruthgn/bank-marketing-data-set}}, Credit Card Fraud Detection\footnote{\url{https://www.kaggle.com/datasets/mlg-ulb/creditcardfraud}}, and Adult Census Income (ACI)\footnote{\url{https://shap.readthedocs.io/en/latest/generated/shap.datasets.adult.html}}. As previously mentioned, we picked four models used in previous studies showing the best performance~\cite{borisov2022deep, gorishniy2021revisiting, grinsztajn2022tree}, including XGBoost, TabNet, FTT, and Saint. 


\subsection{Environment and System Specification}
All experiments are conducted on a Red Hat Enterprise Linux 9 (int5 5.14.0-427.31.1.el9\_4.x86\_64) system equipped with Intel Xeon Platinum 8360Y and AMD EPYC 9334 CPUs, 1TiB RAM, and an NVIDIA H100 and A100. Python 3.11.3 and Pytorch 2.5.1 were used for all experiments.

\begin{table*}[ht]
\centering
\caption{Overview of the datasets used in experiments (after preprocessing).} 
\label{table:DatasetOverview}
\footnotesize
\resizebox{0.9\textwidth}{!}{%
\begin{tabular}{lccccc}
\toprule
                     & \textbf{ACI} & \textbf{BM} & \textbf{CovType} & \textbf{Credit Card} &
                     \textbf{HIGGS}\\ \midrule
Samples              & 32561         & 11162        & 588892   & 284807 & 11000000    \\
Numerical features   & 5             & 6            & 10       & 30     & 28    \\
Categorical features & 7             & 9            & 44       & 0      & 0   \\
Num. of classes      & 2             & 2            & 7        & 2      & 2    \\
Class Ratio        & (76\%, 24\%) & (53\%, 47\%) & (36\%, 49\%, 6\%, 0.6\%, 1.7\%, 3\%, 3.7\%)     & (99.8\%, 0.2\%) & (47\%, 53\%) \\ \bottomrule
\end{tabular}%
}
\end{table*}


\subsection{Evaluation Metrics}

To assess the effectiveness and stealthiness of backdoor attacks, we utilize the following metrics:
\begin{compactenum}
    \item \textbf{Attack Success Rate (ASR):} This metric quantifies the proportion of triggered inputs that the model classifies as the target label \( y_t \):
    \begin{equation*}
        ASR = \frac{1}{m} \sum_{j=1}^{m} \mathbb{I}\left( \mathcal{F}_{\theta^*}(\hat{\mathbf{x}}_j) = y_t \right),
    \end{equation*}
    where \( \mathcal{F}_{\theta^*} \) denotes the backdoored model, \( \mathbb{I} \) is the indicator function, and \( m \) represents the number of poisoned inputs. A higher ASR signifies a more effective backdoor attack.
    
    \item \textbf{Clean Data Accuracy (CDA):} This metric measures the model's accuracy on clean (unpoisoned) inputs and is defined as:
    \begin{equation*}
        CDA = \frac{1}{n-m} \sum_{i=1}^{n-m} \mathbb{I}\left( \mathcal{F}_{\theta^*}(\mathbf{x}_i) = y_i \right),
    \end{equation*}
    where \( n \) is the total number of inputs, and \( n-m \) represents the count of clean inputs. CDA is compared against the Benign Accuracy (BA) of a clean model to determine whether the backdoor compromises the model's performance on legitimate data. A high CDA indicates minimal impact on the model's clean data performance, ensuring its stealthiness~\cite{koffas2023systematic}.
\end{compactenum}

\subsection{Benign Accuracies}

We report the BA values in Tables~\ref{tab:BA} and~\ref{tab:clean-results}. Table~\ref{tab:BA} demonstrates the BA values for pure numeric datasets HIGGS and Credit Card, and Table~\ref{tab:clean-results} includes the results for datasets containing both categorical and numerical features. All results are aligned with the accuracy values reported in previous studies and benchmarks~\cite{li2022backdoor, grinsztajn2022tree, gorishniy2021revisiting}.

\begin{table}[ht]
    \centering
    \caption{BA results (\%) for datasets containing only numerical columns.}
    \resizebox{0.5\columnwidth}{!}{%
    \begin{tabular}{lcccc}
        \toprule
        & {Credit Card} & {HIGGS} \\
        \midrule
        {TabNet} & 99.87 & 77.85 \\
        {Saint} & 99.94 & 79.89\\
        {FTT}  & 99.95 & 79.44\\
        {XGBoost} & 99.95 & 76.23\\ 
        \bottomrule
    \end{tabular}%
    }
    \label{tab:BA} 
\end{table}

\begin{table}
\centering
\caption{BA results (\%) for different encoding methods}
\label{tab:clean-results}
\resizebox{0.75\columnwidth}{!}{%
\begin{tabular}{@{}lllll@{}}
\toprule
\multirow{2}{*}{Dataset} & \multirow{2}{*}{Model}   & \multicolumn{3}{c}{Method} \\
                         &         & OHE       & Ordinal   & \textbf{Ours} \\ \midrule
\multirow{4}{*}{ACI}     & XGBoost & 85.905 & 85.183 & 85.398 \\
                         & TabNet  & 85.015 & 85.521 & 85.460 \\
                         & FTT     & 86.396 & 86.396 & 86.089 \\
                         & Saint   & 86.366 & 86.366 & 86.320 \\ \midrule
\multirow{4}{*}{BM}      & XGBoost & 85.804 & 85.266 & 85.535 \\
                         & TabNet  & 84.326 & 81.818 & 82.983 \\
                         & FTT     & 83.565 & 87.013 & 82.176 \\
                         & Saint   & 84.595 & 85.446 & 84.998 \\ \midrule
\multirow{4}{*}{CovType} & XGBoost & 96.499 & 96.510 & 96.510 \\
                         & TabNet  & 94.875 & 94.563 & 94.482 \\
                         & FTT     & 96.176 & 96.302 & 96.116 \\
                         & Saint   & 96.854 & 96.947 & 96.753 \\ \midrule
\end{tabular}
}
\end{table}

\subsection{Attack Results}

For our primary experiments, we perform CatBack with \(\mu = 1\), \(\lambda = 0.1\), \(\beta = 0.1\), and \(\epsilon = [0.01, 0.02, 0.05, 0.1]\) with all the target classes possible within a chosen dataset.\footnote{Full experiments results, including the effect of choosing different \(\mu\) values, are provided in Appendix~\ref{sec:attack_plots}.}

For the black box scenario, we choose the XGBoost as our target model. XGBoost is a decision tree model. The attacker cannot optimize the gradient-based perturbation in CatBack using XGBoost. Thus, it can be a good candidate for a black box attack. First, we craft our universal perturbation using another model to which we have white box access. In our experiments, we chose TabNet and optimized the perturbation so that it could successfully change TabNet's prediction to the target class. Once we craft our perturbation, we poison the training set by applying it to chosen samples and then train the XGBoost on the poisoned dataset. Finally, we measure CatBack's performance on the backdoored model using the same perturbation.

Table~\ref{tab:exp-results} shows the attack performance with the selected hyperparameters. 
For all results, the CDA values remain similar to their BA and are thus unaffected by CatBack. This is highly in favor of the attacker remaining stealthy with respect to the clean accuracy drop.
With just \(1\%\) poisoning rate, CatBack can achieve more than \(90\%\) ASR in most cases. The results, though, depend on the dataset and target label. The \(\mu\) hyperparameter extensively affects the trigger optimization overhead. This might not be noticeable for regular datasets where it takes seconds for the trigger to be obtained, but for large datasets like HIGGS (where, in our worst-case report, just the regular clean training of Saint might take hours), it is highly evident. 
As we show in Section~\ref{sec:attack_plots}, \(\mu\) also affects ASR for datasets with fewer samples (e.g., BM). With higher values for \(\mu\), more samples distant to the target class manifold get involved in the perturbation optimization process, thus making it more general and effective. This makes it easier for the perturbation to shift the distribution toward the target class.

On the other hand, bigger $\mu$ increases the chance of backdoor detection, as we see later. The only exception we detected in experiments was the results for TabNet on the Credit Card dataset on target label 1 (see Figure~\ref{fig:exps-models-credit}). Although this did not happen with other models, we assume such results come from the dataset being heavily unbalanced (only around 0.1\% of samples in the whole dataset belong to class number 1), making it difficult for a universal perturbation to perform perfectly on all random samples.

Another interesting observation is the results of a black box attack. We can notice that CatBack performs the same or even better in most cases on XGBoost than other models like TabNet. These observations suggest that CatBack can also be a very powerful threat in scenarios in which the poisoned dataset is provided to a third party (e.g., outsourced training), even though the attacker is unaware of the target model.
The detailed results for all experiments regarding CatBack are provided in Figures~\ref{fig:exps-models-aci} to~\ref{fig:exps-models-higgs} and discussed in Appendix~\ref{sec:attack_plots}.

\begin{table}
\centering
\caption{CDA and ASR (CDA/ASR) for $\mu = 1$.}
\label{tab:exp-results}
\resizebox{\columnwidth}{!}{%
\begin{tabular}{@{}lllllll@{}}
\toprule
\multirow{2}{*}{Dataset} & \multirow{2}{*}{Model} & \multirow{2}{*}{\begin{tabular}[c]{@{}l@{}}Target \\ Label\end{tabular}} & \multicolumn{4}{c}{\(\epsilon\)} \\ \cmidrule(l){4-7} 
                             &                          &   & 0.01         & 0.02         & 0.05         & 0.1          \\ \cmidrule(r){1-7}
\multirow{8}{*}{ACI}      & \multirow{2}{*}{FTT}     & 0 & 86.46/96.85 & 86.09/97.10 & 86.20/98.17 & 85.94/98.96 \\
                          &                          & 1 & 86.34/98.50 & 86.35/99.62 & 85.94/99.77 & 86.10/100.0 \\
                          & \multirow{2}{*}{Saint}   & 0 & 85.91/90.99 & 86.20/98.11 & 85.55/99.55 & 85.95/99.91 \\
                          &                          & 1 & 85.95/92.35 & 85.74/93.38 & 85.51/78.84 & 85.52/95.55 \\
                          & \multirow{2}{*}{TabNet}  & 0 & 85.67/97.39 & 85.40/97.62 & 84.83/96.53 & 84.25/98.42 \\
                          &                          & 1 & 85.37/82.60 & 86.04/97.11 & 85.06/100.0 & 84.78/98.27 \\
                          & \multirow{2}{*}{XGBoost} & 0 & 85.09/99.11 & 85.26/99.05 & 85.24/99.74 & 85.32/99.91 \\
                          &                          & 1 & 85.41/98.50 & 85.20/99.51 & 85.28/99.91 & 85.44/99.94 \\ \hline
\multirow{8}{*}{BM}       & \multirow{2}{*}{FTT}     & 0 & 83.70/87.33 & 86.52/95.75 & 86.25/99.96 & 86.30/100.0 \\
                          &                          & 1 & 86.61/98.52 & 86.43/99.91 & 86.57/100.0 & 85.49/100.0 \\
                          & \multirow{2}{*}{Saint}   & 0 & 85.09/96.33 & 84.51/98.88 & 84.68/100.0 & 85.85/99.96 \\
                          &                          & 1 & 85.36/10.00 & 85.85/100.0 & 85.31/100.0 & 84.51/100.0 \\
                          & \multirow{2}{*}{TabNet}  & 0 & 83.88/91.27 & 84.06/89.57 & 83.48/96.78 & 83.16/98.30 \\
                          &                          & 1 & 83.74/84.73 & 82.22/88.22 & 82.80/97.49 & 79.94/96.95 \\
                          & \multirow{2}{*}{XGBoost} & 0 & 84.68/93.55 & 84.51/95.39 & 84.73/97.85 & 84.55/99.46 \\
                          &                          & 1 & 84.86/95.21 & 84.51/97.27 & 84.95/99.24 & 84.55/99.96 \\ \hline
\multirow{28}{*}{CovType} & \multirow{7}{*}{FTT}     & 0 & 95.76/99.98 & 96.23/100.0 & 96.21/100.0 & 95.90/100.0 \\ 
                          &                          & 1 & 96.30/100.0 & 96.28/100.0 & 96.23/100.0 & 96.17/100.0 \\
                          &                          & 2 & 96.31/99.99 & 96.15/100.0 & 96.29/100.0 & 96.03/100.0 \\
                          &                          & 3 & 96.27/99.99 & 96.23/99.99 & 96.21/100.0 & 96.15/100.0 \\
                          &                          & 4 & 96.16/99.98 & 96.20/100.0 & 96.15/100.0 & 96.03/100.0 \\
                          &                          & 5 & 96.30/100.0 & 96.20/100.0 & 96.01/100.0 & 96.15/100.0 \\
                          &                          & 6 & 97.12/99.99 & 96.86/100.0 & 97.09/100.0 & 97.01/100.0 \\
                          & \multirow{7}{*}{Saint}   & 0 & 96.76/100.0 & 96.88/99.96 & 96.83/100.0 & 96.58/99.99 \\
                          &                          & 1 & 96.81/99.98 & 96.73/99.97 & 96.79/99.70 & 96.65/100.0 \\
                          &                          & 2 & 96.83/99.98 & 96.75/99.85 & 96.74/100.0 & 96.61/99.91 \\
                          &                          & 3 & 96.83/99.97 & 96.84/99.84 & 96.77/99.85 & 96.49/99.87 \\
                          &                          & 4 & 96.85/99.92 & 96.85/99.99 & 96.72/100.0 & 96.60/97.64 \\
                          &                          & 5 & 96.93/99.97 & 96.94/100.0 & 96.89/99.98 & 96.54/99.95 \\
                          &                          & 6 & 98.18/99.97 & 98.14/99.96 & 98.16/100.0 & 97.60/99.99 \\
                          & \multirow{7}{*}{TabNet}  & 0 & 94.13/99.93 & 94.72/99.97 & 94.36/100.0 & 93.73/100.0 \\
                          &                          & 1 & 94.10/99.90 & 94.54/99.99 & 94.37/100.0 & 94.38/100.0 \\
                          &                          & 2 & 93.32/99.97 & 92.80/100.0 & 94.47/100.0 & 94.66/100.0 \\
                          &                          & 3 & 94.53/99.95 & 91.07/100.0 & 94.21/100.0 & 94.22/99.98 \\
                          &                          & 4 & 92.18/99.88 & 94.39/99.99 & 94.59/100.0 & 93.32/100.0 \\
                          &                          & 5 & 93.20/99.97 & 94.75/99.91 & 94.70/100.0 & 94.75/100.0 \\
                          &                          & 6 & 95.24/99.96 & 95.64/99.96 & 95.41/99.99 & 95.58/100.0 \\
                          & \multirow{7}{*}{XGBoost} & 0 & 96.51/99.98 & 96.49/100.0 & 96.44/100.0 & 96.32/100.0 \\
                          &                          & 1 & 96.50/100.0 & 96.51/100.0 & 96.42/100.0 & 96.38/100.0 \\
                          &                          & 2 & 96.55/99.99 & 96.49/100.0 & 96.42/100.0 & 96.38/100.0 \\
                          &                          & 3 & 96.52/100.0 & 96.47/100.0 & 96.47/100.0 & 96.35/100.0 \\
                          &                          & 4 & 96.51/99.99 & 96.47/99.99 & 96.46/100.0 & 96.40/100.0 \\
                          &                          & 5 & 96.54/99.99 & 96.54/100.0 & 96.43/100.0 & 96.39/100.0 \\
                          &                          & 6 & 98.73/99.99 & 98.69/100.0 & 98.71/100.0 & 98.56/100.0 \\ \hline
\multirow{8}{*}{Credit}   & \multirow{2}{*}{FTT}     & 0 & 99.95/99.86 & 99.95/99.86 & 99.92/99.86 & 99.94/99.87 \\
                          &                          & 1 & 99.95/99.99 & 99.95/99.98 & 99.95/99.99 & 99.95/99.98 \\
                          & \multirow{2}{*}{Saint}   & 0 & 99.95/99.87 & 99.96/99.85 & 99.96/99.86 & 99.96/99.86 \\
                          &                          & 1 & 99.95/99.90 & 99.94/99.87 & 99.95/99.99 & 99.96/99.98 \\
                          & \multirow{2}{*}{TabNet}  & 0 & 99.85/99.98 & 99.93/99.85 & 99.83/100.0 & 99.92/99.82 \\
                          &                          & 1 & 99.93/99.71 & 96.93/95.15 & 99.93/99.95 & 99.91/86.18 \\
                          & \multirow{2}{*}{XGBoost} & 0 & 99.96/99.87 & 99.96/99.87 & 99.96/99.88 & 99.95/99.90 \\
                          &                          & 1 & 99.96/99.93 & 99.96/99.97 & 99.96/99.95 & 99.96/99.99 \\ \hline
\multirow{8}{*}{Higgs}    & \multirow{2}{*}{FTT}     & 0 & 83.98/99.99  & 83.95/100.0  & 83.87/100.0  & 83.59/100.0 \\
                          &                          & 1 & 83.89/100.0  & 83.96/100.0  & 83.95/100.0  & 83.81/100.0 \\
                          & \multirow{2}{*}{Saint}   & 0 & 80.70/95.00  & 80.74/99.13  & 80.70/99.35  & 80.63/100.0 \\
                          &                          & 1 & 80.67/94.94  & 80.64/99.95  & 80.59/95.58  & 80.48/99.95 \\
                          & \multirow{2}{*}{TabNet}  & 0 & 77.70/99.98  & 78.47/100.0  & 78.18/100.0  & 78.49/100.0 \\
                          &                          & 1 & 78.21/100.0  & 78.48/100.0  & 78.00/99.95  & 78.30/100.0 \\
                          & \multirow{2}{*}{XGBoost} & 0 & 77.38/100.0  & 77.35/100.0  & 77.36/100.0  & 77.35/100.0 \\  
                          &                          & 1 & 77.41/100.0  & 77.34/100.0  & 77.37/100.0  & 77.35/100.0 \\ \cmidrule(l){1-7}  
\end{tabular}
}
\end{table}

\subsection{Baselines}
We have also compared our attack with two baseline attacks. First, we tested Badnets~\cite{gu2019badnets}. To transfer Badnets to the tabular data, we applied the equivalent of an $8\times8$ square trigger in an MNIST image on each poisoned sample. In particular, the perturbation is 0.08\% of all features (MNIST images are $28\times28$ pixels so $(8\times8)/(28\times28)$ = 0.08). We chose a random value within the valid range for each feature for the trigger. We also ran experiments with Tabdoor~\cite{pleiter2023tabdoor} using the in-bounds method. We did not try the out-of-bounds attack since the trigger uses values that are 10\% higher than the maximum value of the poisoned features, which can be easily spotted as outliers. The same is true for the approach used in DBA, where the authors, to create a poisoned sample, changed the values in 6 features of the LOAN dataset using values larger than the maximum value for each of these features. 

The results from these experiments are shown in Table~\ref{tab:baselines}. We did not include the CDA in the tables, as in all cases the performance of the poisoned models was similar to the clean model's performance. 
According to the tables, our attack performs similarly to the baselines for CovType, Credit Card, and Higgs. In these cases, we see that the attack performance is almost perfect (e.g., close to 100\%) in all datasets, meaning that these settings are the easiest to attack. For this reason, CatBack cannot show significantly better performance than the baselines. 
On the other hand, CatBack performs significantly better for ACI (up to 95\% and 44\% better performance than Tabdoor and Badnets, respectively) and BM (up to 71\% and 16\% better performance for Tabdoor and Badnets) datasets. 
Additionally, our trigger can be applied to any data type (numerical and categorical), while the previous attacks were applied only to numerical data. 

\begin{table}[ht]
\centering
\caption{Comparison of CatBack with baseline attacks. Catback Settings: \(\mu=1.0,\; \epsilon=0.02,\) target\_label=1.}
\begin{tabular}{llcccc}
\toprule
Dataset & Attack & FTT & SAINT & TabNet & XGBoost \\
\midrule
\multirow{3}{*}{ACI} 
& Badnets & 86.79 & 87.76 & 53.02 & 99.98 \\
& Tabdoor & 15.16 & 17.49 & 2.02  & 18.61 \\
& CatBack & 99.62 & 93.38 & 97.11 & 99.51 \\
\midrule
\multirow{3}{*}{BM} 
& Badnets & 96.04 & 95.90 & 72.97 & 99.73 \\
& Tabdoor & 28.84 & 78.10 & 24.23 & 99.31 \\
& CatBack & 99.91 & 100.00 & 88.22 & 97.27 \\
\midrule
\multirow{3}{*}{CovType} 
& Badnets & 100.00 & 100.00 & 100.00 & 100.00 \\
& Tabdoor & 99.96  & 99.98  & 99.77  & 100.00 \\
& CatBack & 100.00 & 99.97  & 99.99  & 100.00 \\
\midrule
\multirow{3}{*}{Credit} 
& Badnets & 99.99 & 99.90 & 99.94 & 100.00 \\
& Tabdoor & 99.85 & 99.77 & 97.26 & 99.97 \\
& CatBack & 99.98 & 99.87 & 95.15 & 99.97 \\
\midrule
\multirow{3}{*}{HIGGS} 
& Badnets & 100.00 & 100.00 & 99.98 & 100.00 \\
& Tabdoor & 100.00 & 93.63 & 99.99 & 100.00 \\
& CatBack & 100.00 & 99.95 & 100.00 & 100.00 \\
\bottomrule
\end{tabular}
\label{tab:baselines}
\end{table}

\subsection{Additional Threat Scenario}
As we mentioned in Section~\ref{sec:attack_methodology}, the attacker reverts the poisoned dataset back to its original form after the poisoning is completed. In this way, the users can download and use the dataset in their own pipelines in its original form. $Revert(.)$ is an extra step of our attack, which could be omitted if our encoding method is used directly for the model training by the users, making our attack more efficient. In this case, our encoding method should not affect the model's performance.

To show that this is a viable scenario, we compared the performance of models trained using our encoding with the performance of models trained using ordinal or categorical encoding. From the results shown in Table~\ref{tab:clean-results}, we see that our encoding method could be used as an alternative method for training models on tabular data without hurting the model's performance. This would make our attack easier, as in that case, the attacker should not revert the dataset back to its original form after the training completes. Additionally, our numerical encoding could facilitate input preprocessing in some cases (e.g., transformers like Saint and FTT) as it does not require an embedding layer as the model's input. Finally, our encoding does not result in an increase in the number of the table's columns, as this is common for methods like one-hot encoding, where each categorical column is converted to several binary columns.

\subsection{Evaluation on Vertex AI}
\label{sec:vertexai}

Cloud computing has become the backbone of modern computing infrastructure, with widespread adoption across industries. Organizations of all sizes increasingly rely on cloud platforms to power machine learning workflows. Google’s Vertex AI, Amazon SageMaker, and Azure Machine Learning are the top three services used by almost every company in the market~\cite{statista2024cloudmarket}. Among them, Vertex AI provides a dedicated framework for tabular data through AutoML~\cite{google_vertex_ai_tabular_training}. AutoML has an automated pipeline and algorithms to train and deploy the model. While specific algorithmic defenses within AutoML are not publicly detailed, they include automated data preprocessing (data cleaning, feature engineering, and outlier detection) and robust training practices (input validation, model armor, monitoring for prediction drift, etc.)~\cite{googlecloud2025}. 

As a real-world use case for our black box scenario, we decided to test whether CatBack can be effective on Google AutoML and see whether AutoML's internal mechanisms can mitigate or prevent our attack. Table~\ref{tab:vertexai} demonstrates the feasibility of such attacks in practice. In particular, in all cases, the attack is almost perfect ($>$ 97\%), and the platform did not provide any warnings about the data or the backdoor. Thus, even though this platform includes tools that perform automated data processing, the attack has not been spotted. 
Such an attack can affect the performance of deployed models in production through these platforms (highly likely for many instances, like the financial sector). 

\begin{table}[ht]
\centering
\caption{CatBack Performance on Google AutoML. The attack settings: \(\mu=1.0,\; \epsilon=0.02,\) target\_label=1.}
\begin{tabular}{lccc}
\toprule
Dataset   & BA (\%)       & CDA (\%)   & ASR (\%) \\ \midrule
ACI   & 87.31          & 87.21        & 98.11  \\
Credit Card     & 99.95           & 99.97         & 100.0 \\
CovType   & 96.19            & 97.33         & 100.0 \\
HIGGS  & 76.97            & 76.63         & 100.0 \\
BM  & 86.18            & 87.86         & 96.99 \\

\bottomrule
\end{tabular}
\label{tab:vertexai}
\end{table}

\section{Evaluation Against Defenses}
\label{sec:defenses}

In this section, we investigate how Catback performs against benchmark defenses. We have evaluated our attack against various defense types (including backdoor detection and backdoor removal methods). We also evaluated the poisoned dataset on popular outlier detection methods.  

\subsection{Spectral Signatures}
To evaluate CatBack's stealthiness, we implemented Spectral Signatures~\cite{tran2018spectral} on the poisoned datasets. Using the default settings from the paper, we assume the detector already knows \(\epsilon\) and pinpoints \(1.5 \times \epsilon\) of samples with the highest score as suspicious to remove them from the dataset. We experiment with different \(\mu\) and \(\epsilon\) values. As expected, there is an inverse relationship between \(\epsilon\) and the stealthiness of the attack, as lower poisoning rates bypass the detection mechanism. 
Nonetheless, we observe that \(\mu\) is a more important hyperparameter.

The higher \(\mu\) values mean the universal perturbation is more general and should shift more random samples toward the decision manifold of the target class. On the other hand, smaller \(\mu\) means choosing the samples closer to the decision manifold, which leads to a smaller perturbation needed for the shift. Thus, the poisoned samples are stealthier. For some datasets like BM, ASR is affected by \(\mu\) value, especially when the poisoning rate is small (Figure~\ref{fig:exps-models-bm}). As the BM dataset is the smallest dataset in our experiments, we believe that there are fewer samples very close to the decision boundary, as it can be easier for the model to create a decision boundary in a sparse feature space. When we poison a dataset, we randomly select data from the training set, and by using a small poisoning rate in this case, we may select samples that are further away from the decision boundary. Thus, when $\mu$ is small and the trigger magnitude is also small, the trigger fails to alter the class of the poisoned data. However, if $\mu$ is large, the trigger magnitude is larger, which increases the ASR even if the samples are originally further away from the decision boundary.
Thus, we expect it to leave a stronger effect on stealthiness as well. 

\begin{figure}[htbp]
    \centering
    \begin{subfigure}[t]{0.48\columnwidth}
        \centering
        \includegraphics[width=\linewidth]{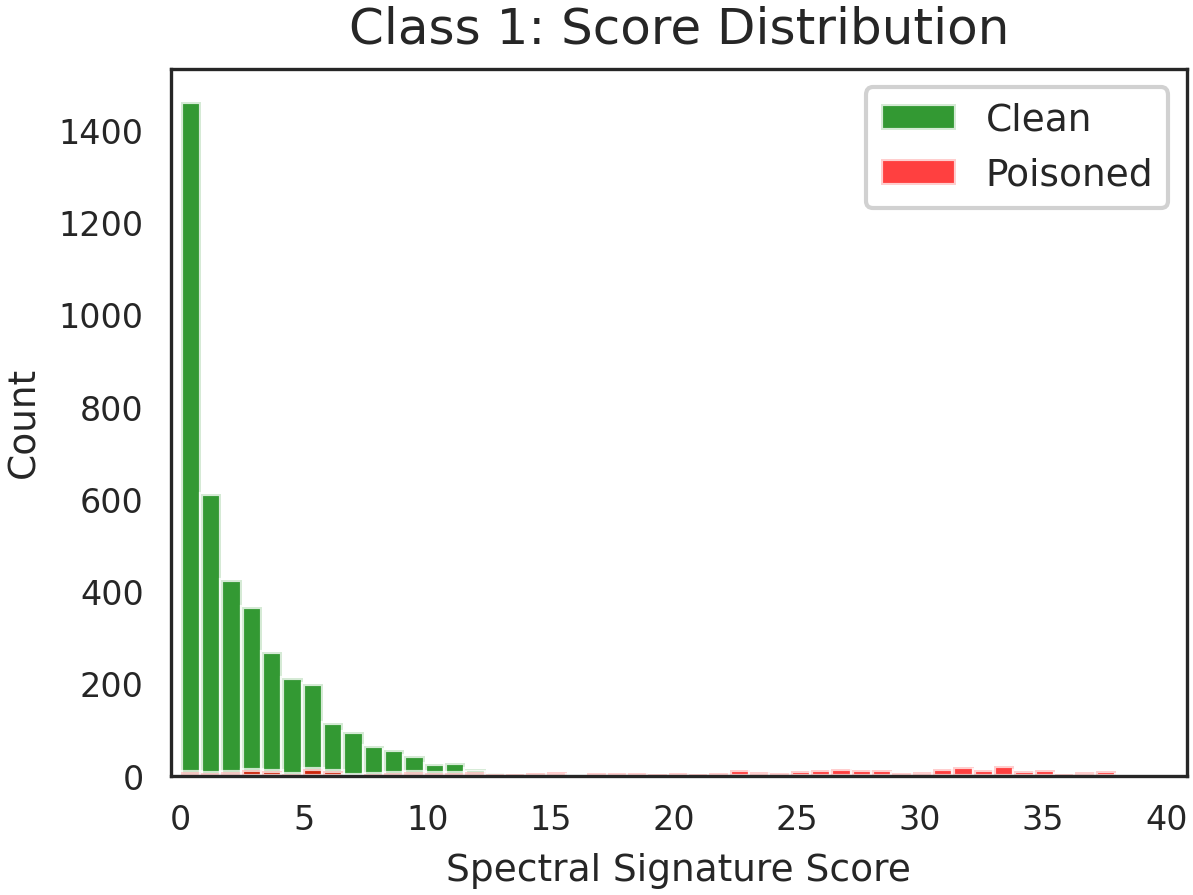}
        \caption{\(\mu=0.5\).}
        \label{fig:mu_value_comparison_ftt_a}
    \end{subfigure}
    \hfill
    \begin{subfigure}[t]{0.48\columnwidth}
        \centering
        \includegraphics[width=\linewidth]{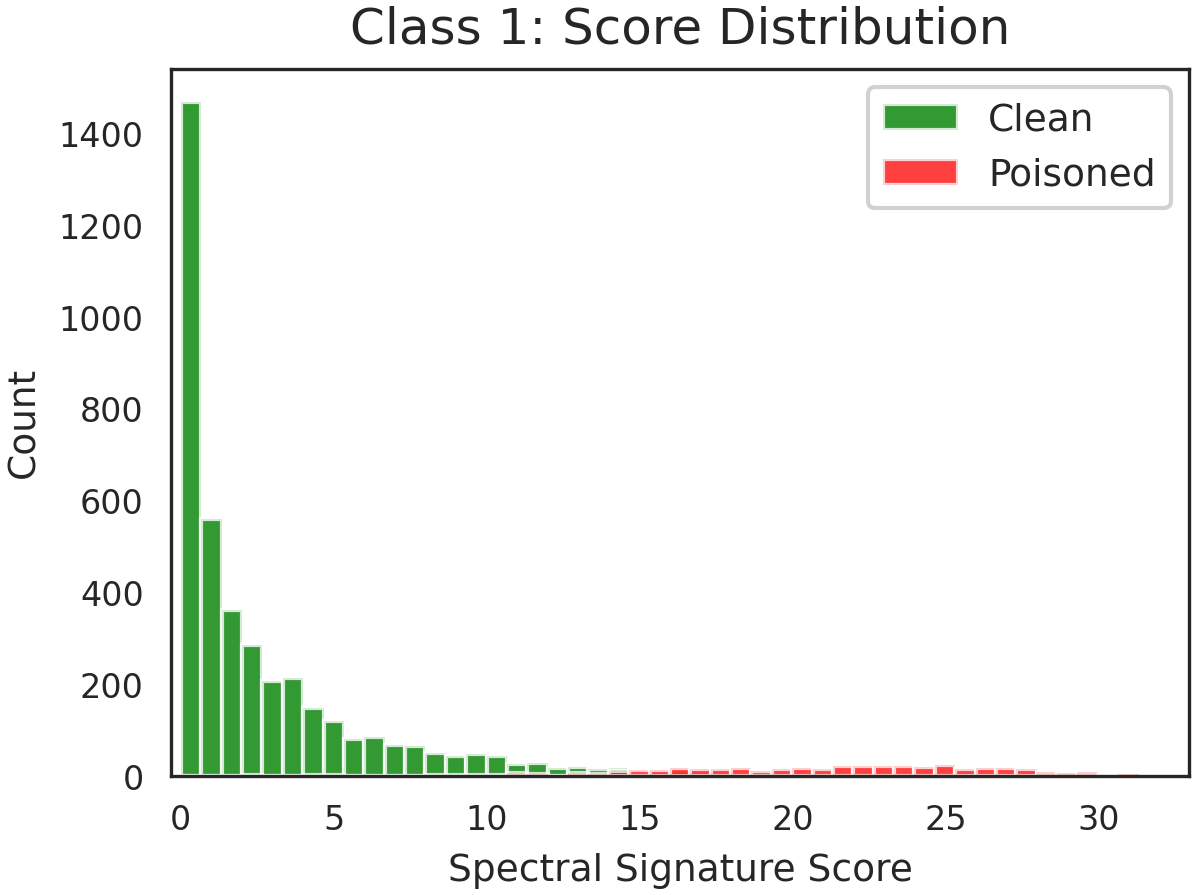}
        \caption{\(\mu=1\).}
        \label{fig:mu_value_comparison_ftt_b}
    \end{subfigure}
    \caption{Spectral Signatures: target class 1, Model: FTT, Dataset: BM, \(\epsilon = 0.05\).}
    \label{fig:mu_value_comparison_ftt_bm}
\end{figure}

\begin{figure}[htbp]
    \centering
    \begin{subfigure}[t]{0.48\columnwidth}
        \centering
        \includegraphics[width=\linewidth]{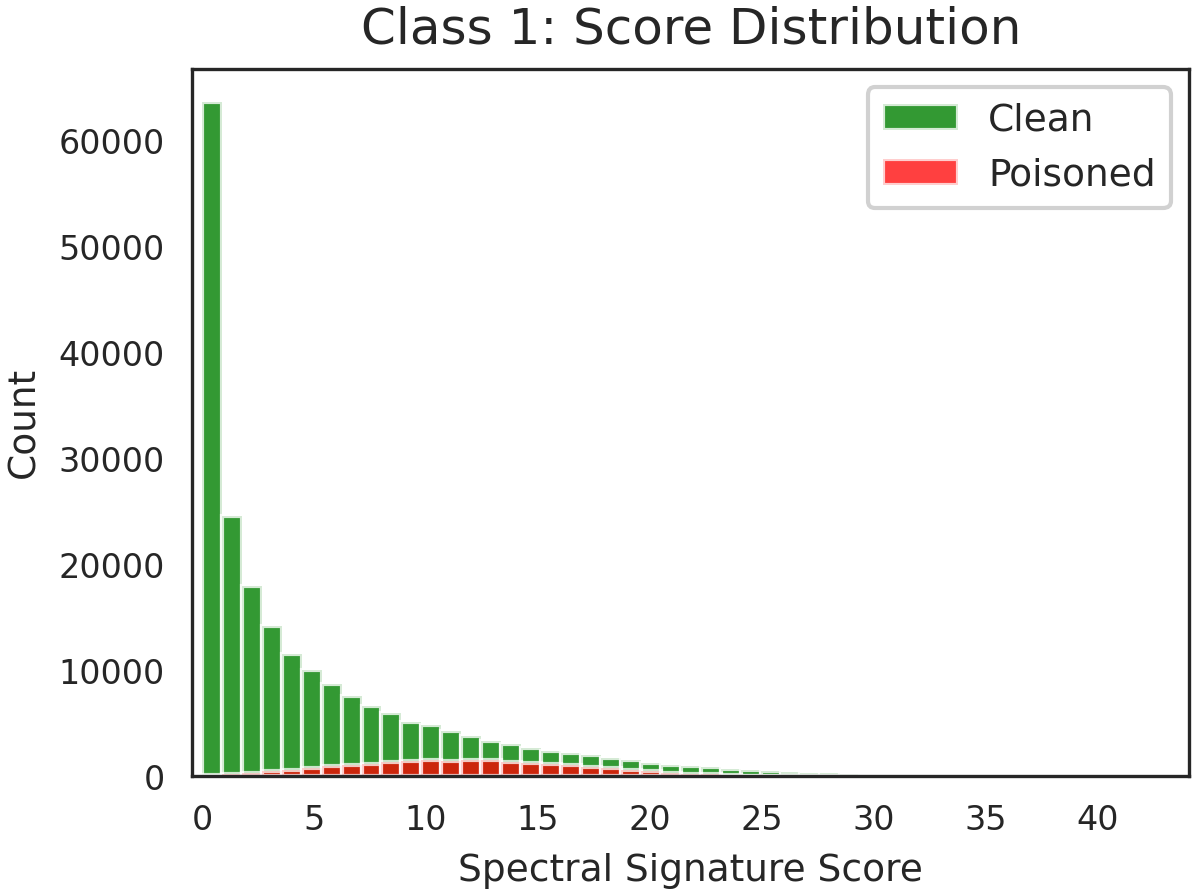}
        \caption{\(\mu=0.5\).}
        \label{fig:mu_value_comparison_ftt_covtype_a}
    \end{subfigure}
    \hfill
    \begin{subfigure}[t]{0.48\columnwidth}
        \centering
        \includegraphics[width=\linewidth]{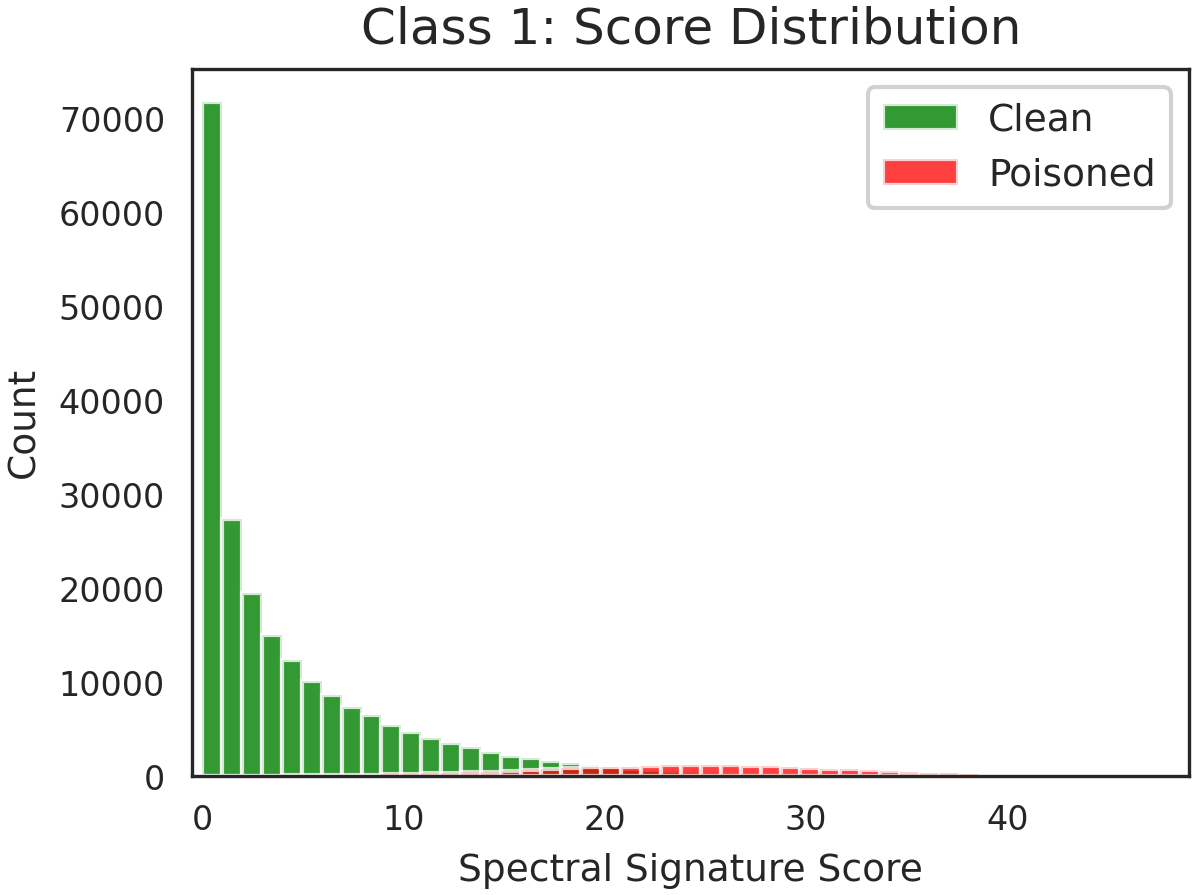}
        \caption{\(\mu=1\).}
        \label{fig:mu_value_comparison_ftt_covtype_b}
    \end{subfigure}
    \caption{Spectral Signatures: target class 1, Model: FTT, Dataset: CovType, \(\epsilon = 0.05\).}
    \label{fig:mu_value_comparison_ftt_covtype}
\end{figure}

We report the results for what be believe to be the least favorable condition for the attacker. Here, we chose the highest \(\mu\) value of 1, \(\epsilon = 0.02\) and target label 1. In this setting, the attack is mostly discoverable for the BM dataset (as we already expected since \(\mu\) had the most influence on ASR for this dataset). Nonetheless, Spectral Signatures fails to remove or detect the poisoned samples in more than \(75\%\) cases. It is worth noting that with other settings, the defense failure rate is even higher, and for brevity, we only report the best defense performance. As an example, we see that by decreasing \(\mu\) value to \(0.5\), it becomes more difficult to remove all poisoned samples for BM (Figure~\ref{fig:mu_value_comparison_ftt_bm}), and it entirely fails to detect poisoned samples for CovType with the FTT model (Figure~\ref{fig:mu_value_comparison_ftt_covtype}).\footnote{Due to space limitations, we include the rest of the figures in the supplementary file in our repository.} 

Moreover, in most cases, the score distributions for non-target classes are the same as the target class, which leads to removing the same number of samples from every non-target class, resulting in a performance reduction of the model on clean data. Previous works that used simple heuristics for the backdoor trigger, such as~\cite{pleiter2023tabdoor}, resulted in more obvious triggers that could be spotted by Spectral Signatures. On the contrary, Spectral Signatures defense is not successful against CatBack, indicating that our attack is stealthy.

\subsection{Neural Cleanse}
We evaluated our attack against Neural Cleanse~\cite{wang2019neural} to measure how detectable the target class is. We adapted the defense for tabular data but adhered to the same hyperparameters and settings from the original paper. Our results show that Neural Cleanse fails to detect the target class. All the anomaly scores we got were below 2 (the threshold), which indicates a complete failure of the method. In fact, most of the scores show that the retrieved mask was around \(0.67 \times MAD\), similar to clean scores. Interestingly, this is the standard deviation of the Laplace distribution used in calculating the $z$-score values for outlier detection, which refers to the normal range of values around the Median (for a $z$-score to be an outlier, the value should be 3 times more than that). In Table~\ref{tab:neural-cleanse-defense}, we show our results for target\_label=1. The results for other target labels are omitted as they demonstrate similar behavior.

\begin{table}[ht]
\centering
\caption{Anomaly Scores for Neural Cleanse. The attack settings: \(\mu=1.0,\; \epsilon=0.02,\) target\_label=1.}
\label{tab:neural-cleanse-defense}
\begin{tabular}{lcc}
\toprule
Model   & Dataset       & Anomaly Score \\
        &               & Backdoored/Clean  \\ \midrule
FTT     & ACI           & 0.6745 / 0.6744        \\
Saint   & ACI           & 0.3699 / 0.6735        \\
TabNet  & ACI           & 0.6694 / 0.6731        \\
FTT     & BM            & 0.6745 / 0.6745       \\
Saint   & BM            & 0.6716 / 0.6730      \\
TabNet  & BM            & 0.6735 / 0.6744       \\
FTT     & CovType       & 0.0033 / 0.0091   \\
Saint   & CovType       & 1.5170 / 0.6742       \\
TabNet  & CovType       & 0.0046 / 1.9131       \\
FTT     & Credit Card   & 0.6745 / 0.6744       \\
Saint   & Credit Card   & 0.6744 / 0.6744       \\
TabNet  & Credit Card   & 0.6644 / 0.6856      \\
FTT     & HIGGS         & 0.6745 / 0.6844      \\
Saint   & HIGGS         & 0.6699 / 0.6838       \\
TabNet  & HIGGS         & 0.6740 / 0.6846       \\
\bottomrule
\end{tabular}
\end{table}

\subsection{Beatrix}
Beatrix~\cite{DBLP:conf/ndss/MaWSXWX23} leverages the second-order statistics of internal feature representations to identify anomalies introduced by backdoor triggers. The mechanism operates by computing Gram matrices from intermediate activations, which capture the correlations among features. Deviations from the expected behavior of clean samples are quantified using the median and the median absolute deviation (MAD). These deviations are further aggregated through a Kernel Maximum Mean Discrepancy (KMMD) metric and standardized to yield an anomaly score, denoted as $J^{*}$. High values of $J^{*}$ (i.e., $ln(J^{*}) > 2$) indicate significant discrepancies from normal feature distributions, suggesting the potential presence of a backdoor. We evaluated our attack against Beatrix. The results indicate that Beatrix fails to detect the presence of Catback. For most cases, we got \(ln(J^*)\approx-0.39\), which indicates the proximity of the poisoned Gram Matrix to the median. Table~\ref{tab:beatrix-defense} showcases our results for target\_label = 1.

\begin{table}[ht]
\centering
\caption{Score results for Beatrix. The attack settings: \(\mu=1.0,\; \epsilon=0.02,\) target\_label=1.}
\label{tab:beatrix-defense}
\begin{tabular}{lccc}
\toprule
Model   & Dataset       & KMMD      & $\ln(J^*)$ \\ \midrule
FTT     & ACI           & 14.6131   & -0.3938 \\
Saint   & ACI           & 74.0251   & -0.3938 \\
TabNet  & ACI           & 43.0890   & -0.3939 \\
FTT     & BM            & 19.5594   & -0.3938 \\
Saint   & BM            & 28.0811   & -0.3938 \\
TabNet  & BM            & 51.3612   & -0.3938 \\
FTT     & CovType       & 3.7468    & -0.3938 \\
Saint   & CovType       & 13.2158   & -0.0416 \\
TabNet  & CovType       & 24.1161   & 0.1113  \\
FTT     & Credit Card   & 32.1099   & -0.3938 \\
Saint   & Credit Card   & 0.0001    & -9.2103 \\
TabNet  & Credit Card   & 39.4631   & -0.3938 \\
FTT     & HIGGS         & 18.4841   & -0.3938 \\
Saint   & HIGGS         & 24.2661   & -0.3938 \\
TabNet  & HIGGS         & 101.8080  & -0.3938 \\
\bottomrule
\end{tabular}
\end{table}

Overall, the results from both Beatrix and Neural Cleanse highlight the high stealthiness of Catback, underscoring the importance of exercising caution when trusting third parties with trained models or ready-to-train datasets.

\subsection{Fine Pruning}
To measure how robust our attack is, we evaluated it against Fine Pruning~\cite{liu2018fine}. We pruned all feed-forward layers (FNN) within the transformers~\cite{gordon2020compressing,clark2019does} (SAINT and FTT) for up to 90\% of neurons and fine-tuned the model for 5 epochs. For SAINT, we also pruned the penultimate layer to make the pruning stronger. Our results show that overall, fine pruning is not effective enough to prevent the attack from fully functioning. For SAINT, although it can restore the accuracy after fine tuning~\cite{lagunas2021block}, ASR also remains considerably high. For FTT, however, Fine Pruning can successfully remove the attack for binary classification tasks with a smaller number of samples. Nevertheless, it fails to remove the backdoor if the dataset is large (e.g., Higgs). We hypothesize that this is because of a higher number of neurons in FNN being involved in learning the representations when facing more complicated datasets. 
With simpler datasets, few neurons can learn the binary manifold like a logistic regression task. This can also be observed when we observe the results for a multiclass dataset (Covtype), where the method fails in both preventing the attack and keeping the CDA high. To confirm this, we conducted extra experiments on another multiclass dataset: Poker Hand\footnote{\url{https://www.kaggle.com/datasets/hosseinah1/poker-game-dataset/data}}, consisting of 10 classes. First, we conduct the attack and evaluate its performance. Table~\ref{tab:attack_results_poker} demonstrates the CatBack results on Poker Hand, with a successful high ASR. Then we perform the Fine Pruning whose results are shown in Table~\ref{tab:fp_results_poker}, and as we see, it likewise fails to mitigate the attack thoroughly.  

\begin{table}[ht]
\centering
\caption{Fine Pruning Results for the pruning\_rate = 0.9. The attack settings: \(\mu=1.0,\; \epsilon=0.02,\) target\_label=1.}
\begin{tabular}{lccc}
\toprule
Model   & Dataset       & FP-CDA (\%)   & FP-ASR (\%) \\ \midrule
Saint   & ACI           & 87.90         & 84.34 \\
FTT     & ACI           & 85.28         & 25.43 \\
Saint   & BM            & 86.94         & 99.96 \\
FTT     & BM            & 80.92         & 23.29 \\
Saint   & CovType       & 90.55         & 65.50 \\
FTT     & CovType       & 79.34         & 98.74 \\
FTT     & Credit Card   & 99.95         & 0.15  \\
Saint   & Credit Card   & 99.95         & 91.49 \\
FTT     & HIGGS         & 77.24         & 63.99 \\
Saint   & HIGGS         & 78.89         & 77.67 \\
\bottomrule
\end{tabular}
\label{tab:fine_pruning_results}
\end{table}

\begin{table}[ht]
\centering
\caption{Attack Results on Poker Dataset. The attack settings: \(\mu=1.0,\; \epsilon=0.02,\) target\_label=1.}
\begin{tabular}{lccc}
\toprule
Model   & BA (\%)       & CDA (\%)   & ASR (\%) \\ \midrule
Saint   & 99.78           & 99.99         & 100 \\
FTT     & 99.79           & 99.96         & 99.90 \\
TabNet   & 99.14            & 94.62         & 97.51 \\
XGBoost  & 96.82            & 99.63         & 100 \\

\bottomrule
\end{tabular}
\label{tab:attack_results_poker}
\end{table}

\begin{table}[ht]
\centering
\caption{Fine Pruning Results on Poker Dataset. The attack settings: \(\mu=1.0,\; \epsilon=0.02,\) target\_label=1.}
\begin{tabular}{lcc}
\toprule
Model        & FP-CDA (\%)   & FP-ASR (\%) \\ \midrule
Saint              & 99.96         & 66.41 \\
FTT                & 67.55         & 74.91 \\

\bottomrule
\end{tabular}
\label{tab:fp_results_poker}
\end{table}

\subsection{Evaluation with the Outlier Detection Algorithms}
\label{sec:outlier_detection}

We have also investigated whether the poisoned samples created by CatBack can be detected using outlier detection methods. For this, we chose the two most popular outlier detection algorithms~\cite{hodge2004survey}: DBSCAN and Isolation Forest, which are commonly used in real-world applications, like fraud detection in finance or network intrusion detection~\cite{laskar2021extending}. We applied these two methods to our poisoned datasets, and both failed to detect and separate the poisoned and clean samples correctly. As a showcase, we report two of our results for the ACI (binary) and CovType (multiclass) datasets. Table~\ref{tab:confusion_matrices} demonstrates the confusion matrices for Isolation Forest and DBSCAN, respectively. As we can observe, DBSCAN fails to achieve an F1-score of higher than $0.02$ while Isolation Forest performs even worse by reaching a precision and recall of $0$, meaning not a single poisoned sample can be detected. We omit the results for the remaining datasets as they showed a similar behavior.



\begin{table}[t]
\caption{Confusion matrices for poisoned-sample detection
(rows: Actual, columns: Predicted; $N$=Normal, $P$=Poisoned).
Parameters: $\mu=1.0$, $\epsilon=0.02$, $\textit{target\_label}=1$.}
\label{tab:confusion_matrices}
\centering
\scriptsize
\setlength{\tabcolsep}{2.5pt}
\renewcommand{\arraystretch}{1.05}

\begin{tabular}{@{}c|c||c|c@{}}
\toprule
\multicolumn{2}{c||}{\textbf{Isolation Forest}} &
\multicolumn{2}{c}{\textbf{DBSCAN}} \\
\cmidrule(lr){1-2}\cmidrule(lr){3-4}
\textbf{ACI} & \textbf{Covtype} & \textbf{ACI} & \textbf{Covtype} \\
\midrule
\begin{tabular}{@{}lcc@{}}
 & \multicolumn{2}{c}{\textit{Pred.}}\\[-2pt]
\textit{Act.} & N & P\\
N & 25008 & 520\\
P & 520 & 0
\end{tabular}
&
\begin{tabular}{@{}lcc@{}}
 & \multicolumn{2}{c}{\textit{Pred.}}\\[-2pt]
\textit{Act.} & N & P\\
N & 446217 & 9296\\
P & 9296 & 0
\end{tabular}
&
\begin{tabular}{@{}lcc@{}}
 & \multicolumn{2}{c}{\textit{Pred.}}\\[-2pt]
\textit{Act.} & N & P\\
N & 11324 & 14204\\
P & 426 & 94
\end{tabular}
&
\begin{tabular}{@{}lcc@{}}
 & \multicolumn{2}{c}{\textit{Pred.}}\\[-2pt]
\textit{Act.} & N & P\\
N & 392434 & 63079\\
P & 7173 & 2123
\end{tabular}
\\[-1pt]
\multicolumn{1}{c}{\footnotesize $F_1=0$} &
\multicolumn{1}{c}{\footnotesize $F_1=0$} &
\multicolumn{1}{c}{\footnotesize $F_1\approx0.0127$} &
\multicolumn{1}{c}{\footnotesize $F_1\approx0.0143$} \\
\bottomrule
\end{tabular}

\end{table}

\section{Ablation Studies}
\label{sec:ablation}

\subsection{Partial Access to Training Data}

In our main threat model, we assume the attacker has full access to the complete training dataset. This might help the attacker observe more samples to calculate the trigger. To evaluate CatBack's effectiveness in stricter conditions, we assume a restricted threat model in which the attacker has a partial auxiliary dataset. We left only 10\% of the dataset in the hands of the attacker and repeated the experiments. The results in Table~\ref{tab:10percent} show that for datasets with a sufficient number of samples, the ASR remains almost the same, indicating that the trigger is still being calculated successfully. For smaller datasets like ACI and BM, with fewer samples, there is a noticeable drop in ASR; yet, in most cases, the attack remains effective.

\begin{table}
\centering
\caption{ASR when the attacker controls only 10\% of the data. The attack settings: \(\mu=1.0,\; \epsilon=0.02,\) target\_label=1.}
\label{tab:10percent}
\begin{tabular}{@{}cccccc@{}}
\toprule
       & \multicolumn{5}{c}{Dataset}                   \\
Model  & ACI   & BM    & CovType & Credit Card & HIGGS \\ \midrule
FTT    & 90.63 & 60.55 & 100     & 99.42       & 100   \\
Saint  & 93.35 & 81.33 & 100     & 99.65       & 100   \\
TabNet & 88.15 & 52.44 & 99.99   & 99.39       & 100   \\
XGBoost & 95.35 & 82.69 & 99.99 & 99.87 & 99.51

\\ \bottomrule
\end{tabular}
\end{table}

\subsection{Frequency Mappings}

   Our $\text{Conv}(\cdot)$ mapping is strictly frequency-based. Each category is assigned a real value in $[0, 1]$ according to how often it occurs in the dataset (plus a small $\Delta r$ to break ties). Thus, there is always a one-to-one mapping between frequency-mapped categories and their corresponding values in one-hot encoding or any other type of representation. This assures that our method preserves the same semantics needed by neural networks to learn and distinguish between categories of each feature. However, our method might induce an extra semantic to the domain knowledge for neural networks by representing the occurrence of each category when the network learns its relations with other co-occurred values in other features.
   
\begin{figure}[htpb]
    \centering
    \includegraphics[width=\linewidth]{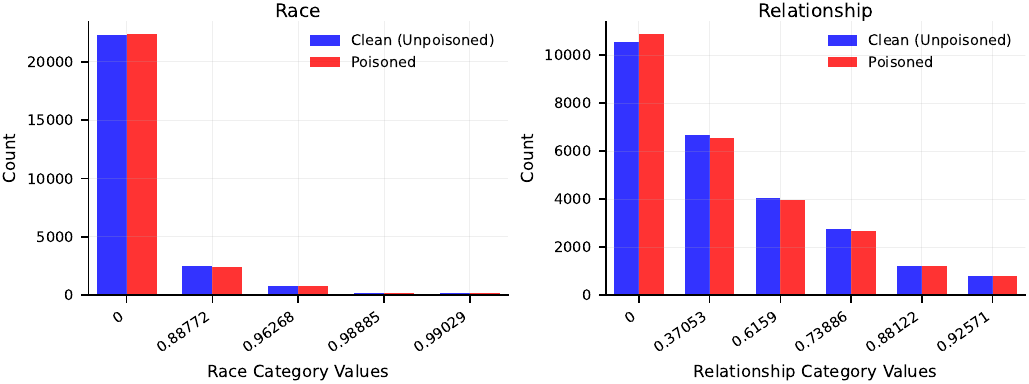}
    \caption{Race and Relationship values before and after poisoning in the ACI dataset.}
    \label{fig:semantic}
\end{figure}

   Another important aspect to notice is that when applying the trigger, the frequency of the categories might change from the original encoding. However, in our attack, the attacker reverts back the dataset to its original values, so the defender would not get suspicious of inconsistency between the frequency mapping values and the frequency of categories in the contaminated dataset, except by accessing the lookup table. Here, the lookup table serves as a secret key, owned and used only by the attacker. One more interesting question is what happens if the user wants to use the raw frequency mappings as the default encoding. Then, the user would still not be able to observe this inconsistency, as the attacker conversion is only for poisoning purposes. After poisoning is done, the poisoned dataset can be converted again to another frequency mapping representation, and the new lookup table is given to the user, but the user never knows there was another lookup table owned by the attacker for poisoning conversion. Even in the worst-case scenario, if the user can obtain the lookup table, the frequency mappings of the clean and poisoned datasets are close, barely raising suspicion in the user. As an example, we demonstrate the Race and Relationship frequency mappings in the ACI dataset in \autoref{fig:semantic}.

\subsection{Performance Under Distribution Shifts}

To conduct further analysis, we designed two experiments for observing attack performance under distribution shifts: 
\begin{enumerate}
    \item In our first experiment, we decided to randomly change the distribution of the entire test set. For doing this, we iterate through all samples, and for each sample, we randomly pick 25\% of all features, whether categorical or numerical (thus, for each sample, the chosen features are different). Then, for each chosen feature, we draw a new value using a Gaussian distribution (within the valid range) and substitute the new value with the old one. Then, we test the model on this newly sampled dataset (containing a new distribution). Table~\ref{tab:combined_distribution_shift} shows that the attack is successful under the presumed condition.

\begin{table}[ht]
\centering
\scriptsize
\caption{Distribution Shift Results on the ACI and Covtype datasets.
Attack settings: \(\mu=1.0,\; \epsilon=0.02,\) target\_label=1.}
\begin{tabular}{llcc}
\toprule
Dataset & Model  & CDA (\%) & ASR (\%) \\ \midrule
\multirow{2}{*}{ACI} 
        & Saint  & 69.80 & 96.26 \\
        & Tabnet & 68.75 & 90.78 \\ \midrule
\multirow{2}{*}{Covtype} 
        & Saint  & 49.70 & 91.61 \\
        & Tabnet & 63.07 & 98.68 \\
\bottomrule
\end{tabular}
\label{tab:combined_distribution_shift}
\end{table}

\item In our second experiment, we implemented concept drift with retraining. Since there should be a pattern in concept drift for the model to learn, we first choose 25\% of the columns randomly from the whole dataset (unlike the previous experiment, this time the chosen columns are fixed for all samples). Then, for each of these columns, if they are categorical, we simply replace them with the least common value for that category in the entire dataset. If the column is numerical, we perform a covariate shift on the values of that column. 

To simulate covariate shift, we apply a sudden distributional change to a subset of samples in selected feature columns. Let $\mathcal{D}_{\text{train}}$ and $\mathcal{D}_{\text{test}}$ denote the training and test datasets respectively, and let $X \in \mathbb{R}^{n \times d}$ represent the feature matrix with $n$ samples and $d$ features.

Let $j \in \{1, \dots, d\}$ denote the index of a selected numerical feature. We denote the values of this feature in the training set as:

$$
X^{(j)}_{\text{train}} = \left\{ x^{(j)}_i \right\}_{i=1}^{n_{\text{train}}}.
$$

We compute the empirical mean and standard deviation of the feature in the training set:

$$
\mu_{\text{train}}^{(j)} = \frac{1}{n_{\text{train}}} \sum_{i=1}^{n_{\text{train}}} x^{(j)}_i, \quad
\sigma_{\text{train}}^{(j)} = \sqrt{ \frac{1}{n_{\text{train}}} \sum_{i=1}^{n_{\text{train}}} \left( x^{(j)}_i - \mu_{\text{train}}^{(j)} \right)^2 }.
$$

We then define transformation parameters for drift:

* A random mean shift factor $\alpha \sim \mathcal{U}(0.1, 0.3)$

* A random scale factor $\beta \sim \mathcal{U}(0.1, 0.3)$

Using these, we compute:

$$
\Delta\mu = \alpha \cdot \sigma_{\text{train}}^{(j)}, \quad
\gamma = 1 + \beta.
$$

The transformation applied to each selected test sample $x^{(j)}$ is:

$$
x^{(j)}_{\text{drifted}} = \left( x^{(j)} + \Delta\mu \right) \cdot \gamma.
$$

This results in a shifted and scaled version of the original feature, effectively altering both the first and second moments of the feature distribution. The operation introduces a sudden covariate shift.

We apply the concept drift on the entire test set and 25\% of the train samples randomly. Then we perform a complete retraining of the model on the drifted training set and evaluate the attack on the test set. The results are demonstrated in Table~\ref{tab:concept_drift}. Results suggest that concept drift and retraining together can impact the ASR; however, this highly depends on the model. 

As a takeaway, we can regard this as a possible solution to mitigate the attack efficacy. Thus, by monitoring and responding to distribution shifts (scheduled retaining), practitioners can remove implanted triggers over time.

\begin{table}[ht]
\centering
\scriptsize
\caption{Concept Drift (with retraining) Results on the ACI and Covtype datasets.
Attack settings: \(\mu=1.0,\; \epsilon=0.02,\) target\_label=1.}
\begin{tabular}{llcc}
\toprule
Dataset & Model  & CDA (\%) & ASR (\%) \\ \midrule
\multirow{2}{*}{ACI} 
        & Saint  & 84.08 & 99.43 \\
        & Tabnet & 85.23 & 19.38 \\ \midrule
\multirow{2}{*}{Covtype} 
        & Saint  & 92.32 & 91.5 \\
        & Tabnet & 92.24 & 24.29 \\
\bottomrule
\end{tabular}
\label{tab:concept_drift}
\end{table}

\end{enumerate}


\section{Limitations and Discussion}
\label{sec:limitations}

Imperceptibility refers to how unnoticeable the trigger is to a human observer or simple statistical checks. Its focus is on the input space (features), and the goal is to make the trigger visually or statistically subtle so it does not raise suspicion~\cite{qin2019imperceptible}.
Stealthiness, however, usually refers to how difficult the attack is to detect by automated defenses or manual inspections, not just in the input, but also in model behavior, the training process, or data patterns. Its focus is on the training process, model internals, and activation patterns~\cite{liu2022hide}.
Its goal is to make the attack difficult for defenders to detect using anomaly detection, activation clustering, or mitigating techniques.
Thus, the attack could be perceptible, yet highly stealthy, hiding from all detection methods~\cite{liu2022hide}.

A variety of domain‐specific metrics have been proposed to quantify imperceptibility in non‐tabular settings:
\begin{itemize}
  \item \textbf{Image Domain:} Metrics based on $L_p$ norms (e.g., $L_2$, $L_\infty$) measure the magnitude of pixel perturbations, while perceptual similarity measures such as Structural Similarity Index (SSIM) and Learned Perceptual Image Patch Similarity (LPIPS) model human visual perception to assess perturbation visibility~\cite{wang2004image, zhang2018unreasonable, wang2024mm}. Frequency‐domain metrics (e.g., DCT distortion) further ensure that the trigger does not introduce atypical spectral components.
  \item \textbf{Text Domain:} Metrics such as word‐embedding distance and perplexity change quantify the semantic drift introduced by triggers. Approaches like BLEU and ROUGE scores, alongside language model‐based fluency measures, evaluate whether inserted or substituted tokens remain coherent in the sentence context.
  \item \textbf{Voice Domain:} Imperceptibility is measured by signal‐to‐noise ratio (SNR), Mel‐cepstral distortion (MCD), and perceptual evaluation of speech quality (PESQ) to ensure that the backdoor trigger (e.g., audio watermark) remains inaudible or indistinguishable from natural background noise.
\end{itemize}

One of the limitations we face during the current study is the lack of an imperceptibility metric in the tabular domain. Although we design our backdoor to be stealthy, there is no accepted measure in the community so that we can calculate how imperceptible the backdoor trigger is in tabular entry. Though designing and proposing a new metric specific to tabular data is out of the scope of this work, we try to illustrate some aspects of a good imperceptibility metric tailored for tabular data, and hope this can provide a good insight for future studies. 
Similar to BlEU or SSIM, we assume that there is a reference sample for the given synthetic sample (here, we consider the original clean sample as the reference and its corresponding backdoored sample as the synthetic). Let us denote the synthetic row as
\[
\mathbf{x} = (x_1, x_2, \dots, x_F)
\]
against a real reference row: 
\[
\mathbf{y} = (y_1, y_2, \dots, y_F)
\]
We assume that there is a tabular imperceptibility score denoted as TIS. We believe that TIS should be a function that is dependent on two important factors: 
\[TIS = f(FLS, IF).\]

\begin{itemize}
    \item Feature-Level Similarity (FLS): A per-feature score that measures how similar each candidate feature is to its reference counterpart.
    \item Interaction Fidelity (IF): A penalty/bonus component that captures the similarity of pairwise (or higher-order) interactions among features.
\end{itemize}

We conjecture that a good TIS should have these values calculated in its process, either explicitly or implicitly. FLS calculation might be just a distance measure (e.g., $L_2$ distance). However, calculating IF might be more complicated and vary from dataset to dataset. For example, we assume there is a tabular dataset consisting of different columns of patient data. Age and BP (bacterial peritonitis) columns might be highly correlated with each other. 
However, this is something that could be known only by specialists or researchers who obtained the correlation function between these features (e.g., through linear regression).
Thus, formulating these correlation functions varies significantly between each dataset and may require the assistance of a specialist in that field.
We leave this as an open question that might be an interesting topic for future studies.

\section{Related Work}
\label{sec:related}

The first backdoor attacks were introduced in~\cite{gu2019badnets,chen2017targeted}. In these works, the authors demonstrated that an attacker can insert a backdoor into a trained model just by altering a few training samples. A large number of different attacks have been introduced since then, most of them targeting computer vision~\cite{li2022backdoor}, but also targeting different domains like graph neural networks~\cite{xi2021graph}, language models~\cite{chen2021badnl}, or speech processing~\cite{zhai2021backdoor}.

Only a few works studied backdoors on tabular data. \cite{xie2019dba} performed distributed backdoor attacks on federated learning. For a backdoor on tabular data, the authors performed a feature importance ranking and chose the 6 least important features. For each of those chosen features, they set their value to an outbound number larger than the maximum value in the dataset. 
\cite{pleiter2023tabdoor} conducted a more comprehensive study on backdoors in tabular data, resulting in the Tabdoor attack. Contrary to~\cite{xie2019dba}, the authors observed less of a relation between feature importance and attack success rate. Since the authors investigated more datasets, they suggested that using the most important features results in a slightly higher attack success rate. 
To improve over~\cite{xie2019dba}, they proposed two different attacks: Out-of-bounds and In-bounds attacks. For Out-of-bounds, they followed the same idea as in~\cite{xie2019dba}, but with fewer features. The authors tried 1, 2, and 3 features and set their values to 10\% higher than the maximum values of each feature in the whole dataset. Since the Out-of-bounds attacks can be easily discovered as outliers,~\cite{pleiter2023tabdoor} performed the In-bounds attack as their main contribution. For In-bounds, they chose the top 3 most important features within the dataset and set their values to the most common value for that feature. As presented in previous sections, CatBack works significantly better than Tabdoor, especially for more complex datasets.

\cite{lv2023data} performs a data-free backdoor attack. They used an approach similar to teacher-student type of training to induce the trigger pattern in the victim model without accessing the dataset. For this, they collect a substitute dataset (which may be irrelevant to the main dataset) and try to implant the trigger value inside the substitute dataset. For tabular data, they chose ACI as their main task dataset and CovType as the substitute. They selected two features of ACI (fnlwgt, sex) with a fixed value and implanted them in the Covtype (with 20\% poisoning rate), while changing the label to target, and then fine-tuned the victim model. Unfortunately, due to a significantly different setup from ours (different models, types of attacks, and different poisoning rates), direct comparison with our work is not possible. 
\cite{10606191} performs a backdoor on Binary click-through rate (CTR) datasets using a Factorization model (Deep-FM). To do so, they conducted a feature importance ranking based on the Deep-FM embedding layer (they consider features with higher average weight values to be more important). Then, they select the top 4 most important features and set them to the least frequent value. However, the authors did not provide the code, and we could not reproduce the results. For this reason, we did not consider this attack in our evaluation.


\section{Conclusions and Future Work}
\label{sec:conclusions}




In this work, we have introduced a targeted backdoor attack on tabular data by crafting a universal trigger pattern that can be added to any input within the domain. By leveraging frequency‑based floating‑point mapping for categorical values, we unify all feature types into a continuous space and enable a single, elastic‑net‑regularized perturbation to reliably shift any sample toward an adversarial target. By enabling the trigger to involve any column of a table, including categorical features, we broaden the attack surface.

We evaluated CatBack across five widely used benchmark datasets and four distinct models, achieving near‑perfect attack success rates (up to 100\%) with negligible impact on clean‑data accuracy, under constraints such as 1\% poisoning rate or only 10\% auxiliary data access. Moreover, our evaluation against state-of-the-art defenses shows that CatBack remains largely undefeated, and it readily transfers to real-world cloud services, such as Google Vertex AI. 

These findings underscore the need for future studies to rethink trust assumptions around tabular pipelines, including the development of custom defenses for tabular data, such as preprocessing-agnostic detection strategies, and incorporating robust certification for mixed-type inputs. Future work could extend CatBack to dynamic, sample‑specific triggers, investigate defenses based on causal feature attribution, and evaluate these threats in federated or privacy‑preserving settings. Ultimately, securing mission‑critical domains such as finance and healthcare will require holistic frameworks that treat tabular ML with the same adversarial scrutiny long applied to vision and language.

\section{Ethical Considerations}

Our work investigates backdoor attacks for tabular data, highlighting how efficient transformation of input data can lead to high-performing and stealthy backdoors. While our goal is to increase awareness about this attack vector, considering the prevalence of tabular data in real-world applications, we acknowledge that disclosing such vulnerabilities could potentially be exploited for malicious purposes. All experiments conducted in this research were designed to avoid exposing sensitive data or causing real-world harm. Evaluations were performed in controlled environments, using publicly available datasets. Despite these precautions, we acknowledge that revealing any new attack carries some risk. Still, we believe that transparent disclosure of such vulnerabilities, combined with responsible communication, benefits the community.

\bibliography{references}
\bibliographystyle{IEEEtran}

\appendices
\section{Example of Our Encoding and Attack Algorithm}
\label{sec:example}

Here, we demonstrate our encoding method through an example. To this end, we consider a categorical feature \( j \) with the following category counts:
\begin{table}[ht]
    \centering
    \begin{tabular}{|c|c|}
        \hline
        \textbf{Category \( v_{jl} \)} & \textbf{Count \( c_{jl} \)} \\
        \hline
        A & 50 \\
        B & 50 \\
        C & 30 \\
        D & 20 \\
        \hline
    \end{tabular}
    \caption{Category Counts for Feature \( j \)}
    \label{tab:category_counts}
\end{table}

\textbf{Primary Mapping:}

\[
\begin{array}{rlrl}
r_{jA} &= \frac{50 - 50}{50 - 1} = 0.0000, & \quad
r_{jB} &= \frac{50 - 50}{50 - 1} = 0.0000 \\[6pt]
r_{jC} &= \frac{50 - 30}{50 - 1} \approx 0.4082, & \quad
r_{jD} &= \frac{50 - 20}{50 - 1} \approx 0.6122
\end{array}
\]

\textbf{Determine \( \Delta r \):}
\begin{itemize}
    \item \(\Delta r_{\text{min}} = 0.204\)
    \item Identify the largest single decimal component in \(\Delta r_{\text{min}}\):
    \begin{itemize}
        \item \(\Delta r_{\text{min}} = 0.204\) has the first non-zero decimal at the first decimal place (\( p = 1 \)).
    \end{itemize}
    \item Set \( \Delta r = 10^{-(p + 1)} = 10^{-2} = 0.01 \).
\end{itemize}

\textbf{Apply Hierarchical Mapping:}
\begin{align*}
r_{jA}' &= 0.0000 + (1 - 1) \times 0.01 = 0.0000 \\
r_{jB}' &= 0.0000 + (2 - 1) \times 0.01 = 0.0100 \\
r_{jC}' &= 0.4082 \\
r_{jD}' &= 0.6122 
\end{align*}

\textbf{Result:}

\begin{table}[ht]
    \centering
    \begin{tabular}{|c|c|c|}
        \hline
        \textbf{Category \( v_{jl} \)} & \textbf{Original \( r_{jl} \)} & \textbf{Updated \( r_{jl}' \)} \\
        \hline
        A & 0.0000 & 0.0000 \\
        B & 0.0000 & 0.0100 \\
        C & 0.4082 & 0.4082 \\
        D & 0.6122 & 0.6122 \\
        \hline
    \end{tabular}
    \caption{Updated Numerical Representation After Hierarchical Mapping}
    \label{tab:updated_mapping}
\end{table}

 \textbf{Lookup Table:}
    
    Referring to Table~\ref{tab:updated_mapping}, the lookup table \( T_j \) for feature \( j \) is constructed as:
    
    \begin{table}[ht]
        \centering
        \begin{tabular}{|c|c|}
            \hline
            \textbf{Numerical Value \( r_{jl}' \)} & \textbf{Category \( v_{jl} \)} \\
            \hline
            0.0000 & A \\
            0.0100 & B \\
            0.4082 & C \\
            0.6122 & D \\
            \hline
        \end{tabular}
        \caption{Lookup Table \( T_j \) for Feature \( j \)}
        \label{tab:lookup_table}
    \end{table}
    
    When an \( r_{jl}' \) value of 0.0100 is encountered during reverse mapping, the corresponding category retrieved from \( T_j \) is B.

\begin{algorithm}[ht]
\footnotesize
\caption{CatBack: Universal Backdoor for Tabular Data}
\label{alg:catback}
\begin{algorithmic}[1]
  \Require $D_{\mathrm{orig}}, t, \mathcal{F}, \mu,\beta,\lambda,\epsilon$
  \Ensure  Backdoored model $\mathcal{F}'$
  \State $D \gets \mathrm{Conv}(D_{\mathrm{orig}})$  \Comment{frequency‑based encoding \& tie‑resolution}
  \State Train $\mathcal{F}$ on $D$
  \State $D_{\mathrm{nt}}\gets\{\,(x,y)\in D \mid y\neq t\}\}$
  \ForAll{$(x_i,y_i)\in D_{\mathrm{nt}}$}
    \State $s_i\gets f_t(x_i)$  \Comment{softmax score for class $t$}
  \EndFor
  \State Sort $D_{\mathrm{nt}}$ by $s_i$ descending
  \State $D_{\mathrm{picked}}\gets$ top $\mu\,|D_{\mathrm{nt}}|$ samples
  \State Initialize $\delta\gets\mathbf{0}\in\mathbb R^d$
  \Repeat
    \ForAll{$(x_i,y_i)\in D_{\mathrm{picked}}$}
      \State $\hat x_i\gets\mathrm{clip}(x_i+\delta)$
      \State $\ell_i\gets -\log f_t(\hat x_i)\;+\;\beta\|\hat x_i-\mathrm{Mode}(D)\|_1\;+\;\lambda\|\hat x_i-\mathrm{Mode}(D)\|_2^2$
    \EndFor
    \State $\delta\gets\delta-\eta\,\nabla_\delta\bigl(\tfrac1{|D_{\mathrm{picked}}|}\sum_i\ell_i\bigr)$
    \State Round categorical dimensions of $\delta$ to nearest valid $r'_{jl}$
  \Until{convergence}
  \State Randomly select $D_{\mathrm{sel}}\subset D$, $|D_{\mathrm{sel}}|=\epsilon\,N$
  \ForAll{$x_i\in D_{\mathrm{sel}}$}
    \State $\hat x_i\gets\mathrm{clip}(x_i+\delta)$, \quad $\hat y_i\gets t$
  \EndFor
  \State $D'\gets \mathrm{Revert}\bigl((D\setminus D_{\mathrm{sel}})\cup\{(\hat x_i,\hat y_i)\}\bigr)$
  \State Preprocess $D'$ (OHE/embeddings/ordinal), train $\mathcal{F}'$ on $D'$
  \Return $\mathcal{F}'$
\end{algorithmic}
\end{algorithm}
\section{Detailed Attack Results}
\label{sec:attack_plots}

\begin{figure*}[htpb]
    \centering
    \includegraphics[width=0.8\textwidth]{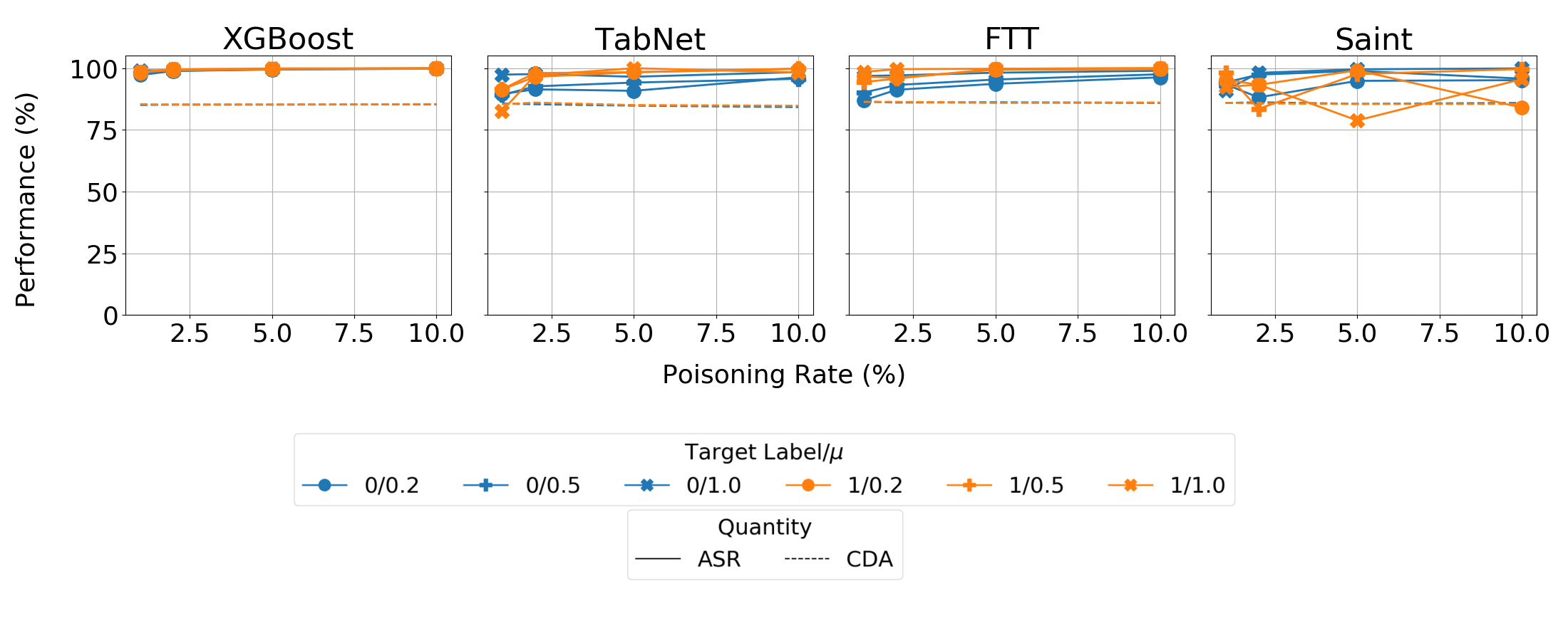}
    \caption{CatBack's ASR and CDA vs. the poisoning rate for different $\mu$ using the ACI dataset.}
    \label{fig:exps-models-aci}
\end{figure*}

\begin{figure*}[htpb]
    \centering
    \includegraphics[width=0.8\textwidth]{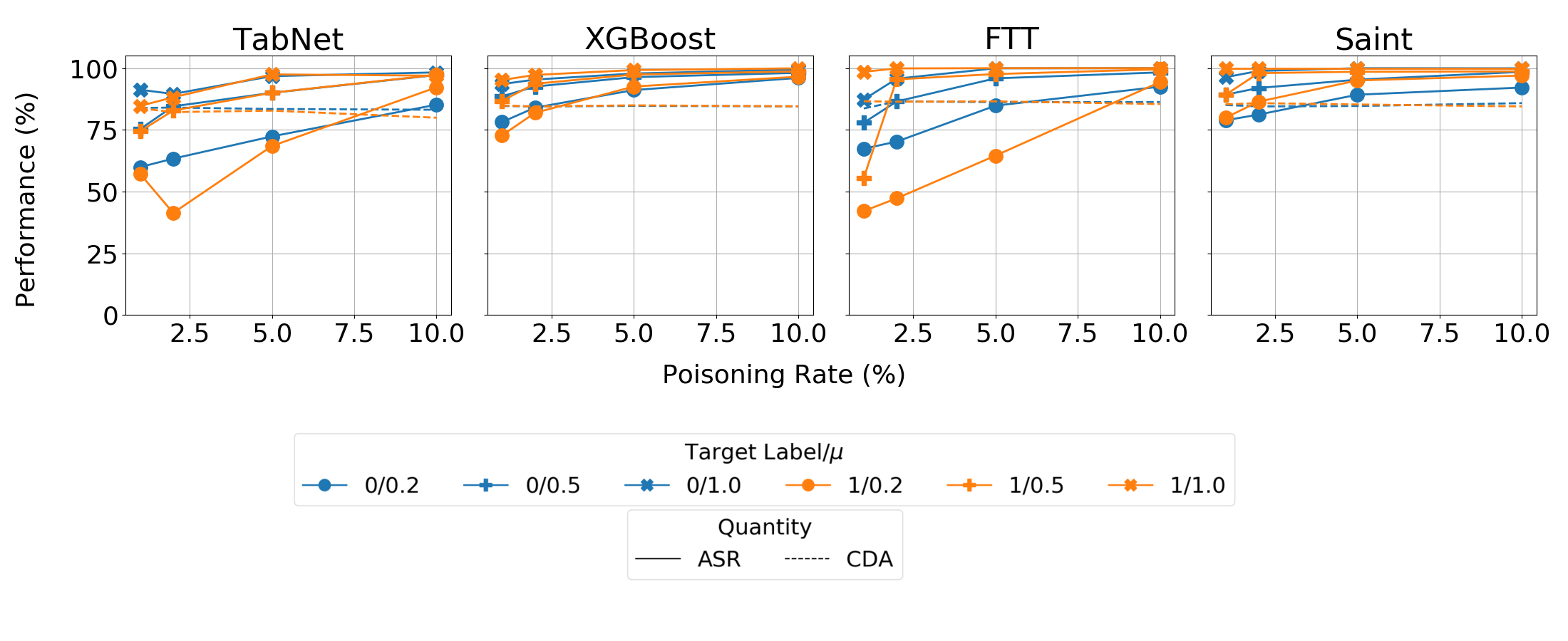}
    \caption{CatBack's ASR and CDA vs. the poisoning rate for different $\mu$ using the BM dataset.}
    \label{fig:exps-models-bm}
\end{figure*}

\begin{figure*}[htpb]
    \centering
    \includegraphics[width=0.8\textwidth]{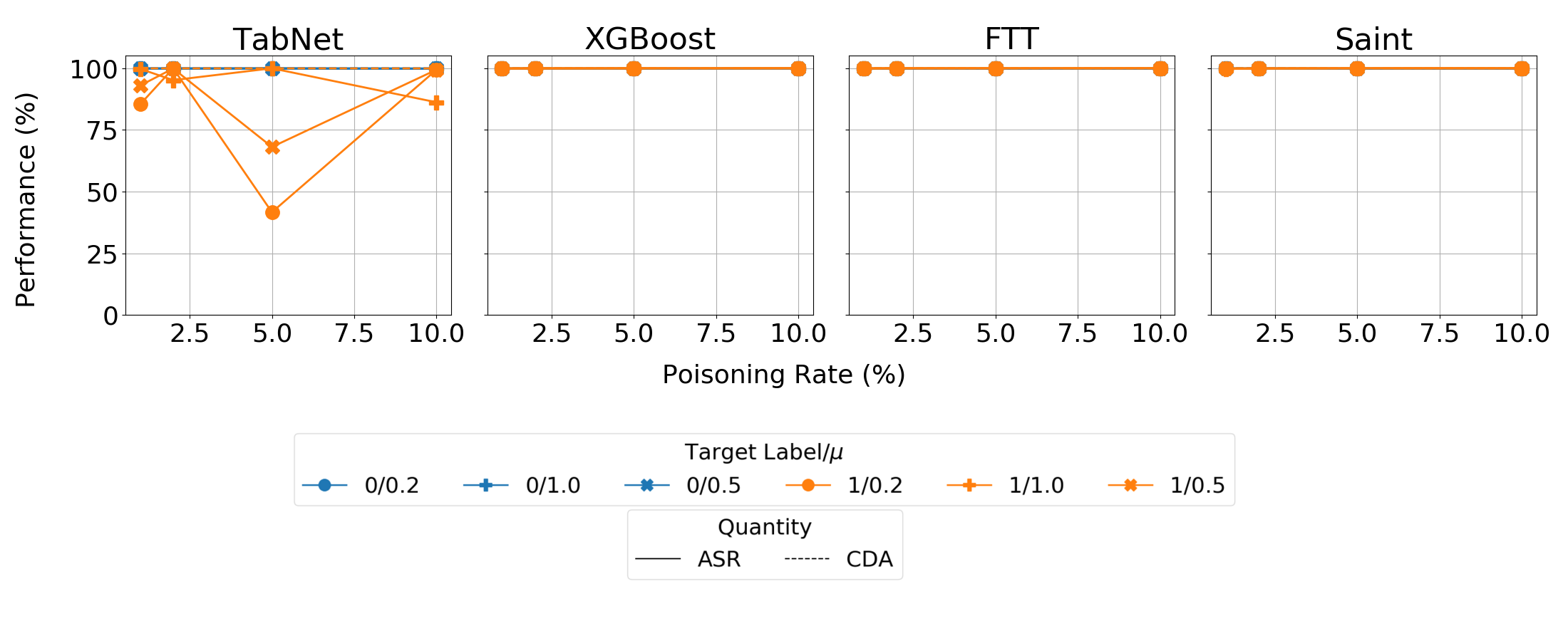}
    \caption{CatBack's ASR and CDA vs. the poisoning rate for different $\mu$ using the Credit Card dataset.}
    \label{fig:exps-models-credit}
\end{figure*}

\begin{figure*}
    \centering
    \includegraphics[width=0.8\textwidth]{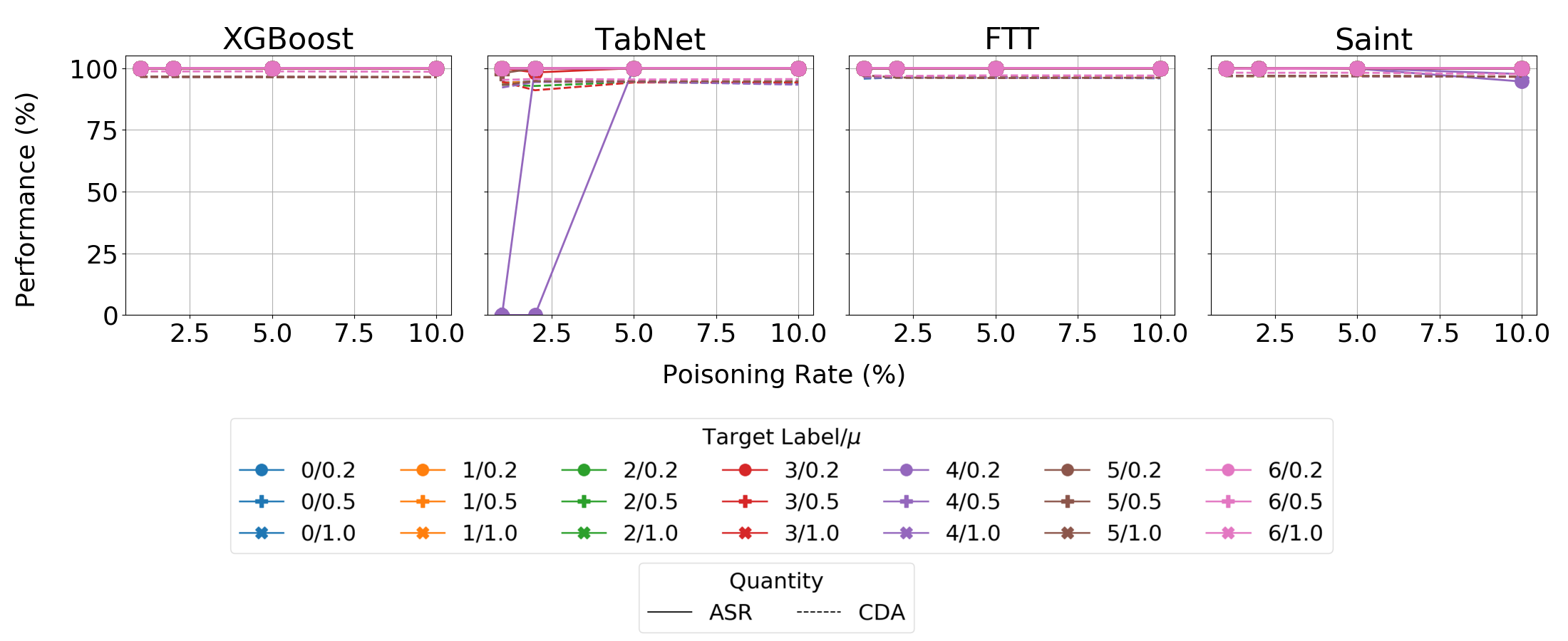}
    \caption{CatBack's ASR and CDA vs. the poisoning rate for different $\mu$ using the Forest Cover Type dataset.}
    \label{fig:exps-models-covtype}
\end{figure*}

\begin{figure*}
    \centering
    \includegraphics[width=0.8\textwidth]{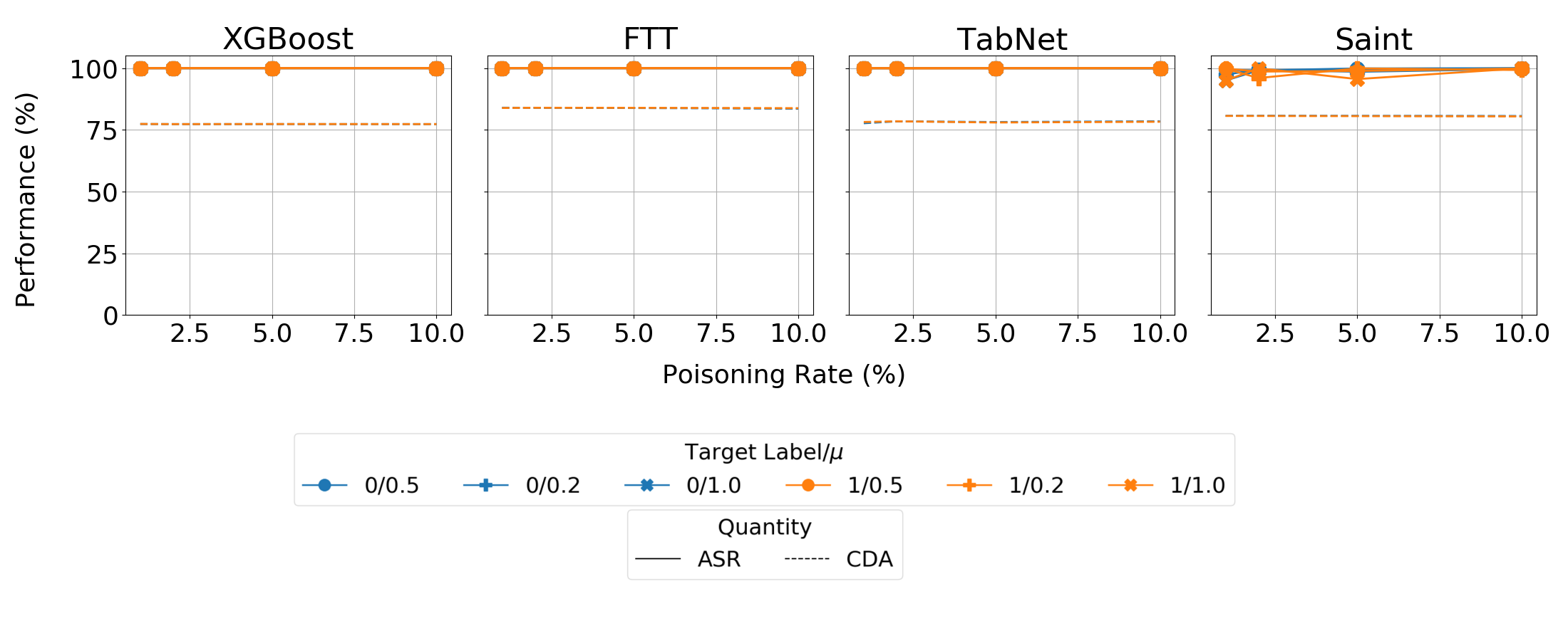}
    \caption{CatBack's ASR and CDA vs. the poisoning rate for different $\mu$ using the HIGGS dataset.}
    \label{fig:exps-models-higgs}
\end{figure*}

In this section, we have included the results from our experiments for all the hyperparameters we evaluated. In particular, in Figures~\ref{fig:exps-models-aci} to~\ref{fig:exps-models-higgs}, we show ASR (solid lines) and CDA (dotted lines) for our experiments for all datasets, models, target labels, and $\mu$ values. In Figure~\ref{fig:exps-models-aci}, we see that our attack achieves a performance of more than 80\% in most cases. Additionally, the performance slightly increases as we increase the poisoning rate. In this case, both the $\mu$ and the target label do not significantly influence CatBack's performance. In Figure~\ref{fig:exps-models-bm}, however, the target label can affect the attack performance significantly in some cases. As BM is a relatively small dataset, we believe that the small imbalance of the two classes could lead to such differences. Class 0 corresponds to 53\% of the dataset, and for this reason, the attack is slightly more effective when the target class is 0.
Additionally, in Figure~\ref{fig:exps-models-bm}, when the FTT is used, we see that $\mu$ also affects CatBack's performance for both target labels, with $\mu=1$ leading to the best performance due to bigger perturbation values.
Similar results were seen for the rest of the datasets as shown in Figures~\ref{fig:exps-models-credit},~\ref{fig:exps-models-covtype}, and~\ref{fig:exps-models-higgs}.

\newpage
\section{Artifact Appendix}
\label{sec:artifact}
%
%

\subsection{Description \& Requirements}
%
%

\subsubsection{How to access}
The evaluator can clone our code from our publicly available repo \url{https://github.com/catback-tabular/catback}.\footnote{We also provide our implementation through the following link (although it is not updateable anymore and may become deprecated): \url{https://doi.org/10.5281/zenodo.17035715}.}

\subsubsection{Hardware dependencies}
\begin{itemize}
    \item \textbf{CPU}: Modern multi-core processor (Intel/AMD x86\_64). We have tested our code on Intel Xeon Platinum 8360Y @ 2.40GHz.
    \item \textbf{RAM}: Minimum 8GB, recommended 16GB+ for larger datasets. Our system has 32GB available.
    \item \textbf{Storage}: At least 16GB free space for datasets and model storage.
    \item \textbf{GPU}: Needed for faster training (CUDA-compatible GPU with 4GB+ VRAM). We have tested NVIDIA A100-SXM4-40GB.
\end{itemize}

\subsubsection{Software dependencies}
\begin{itemize}
    \item \textbf{Operating System}: Linux (we run our experiments on RHEL 9.4 and Ubuntu 24.04)
    \item \textbf{Python}: Version 3.8 or higher. We have run our experiments with Python 3.11.3.
    \item \textbf{CUDA}: version 11.0+ for using GPU acceleration. We have used CUDA 12.4.
    \item \textbf{PyTorch}: Compatible with CUDA version. We have tested PyTorch 2.5.1+cu124.
\end{itemize}

Our code can be run on commodity hardware (standard desktop/laptop). GPU acceleration is optional, but it significantly speeds up training.

\subsubsection{Benchmarks}
As we discuss in our repo's README file, we have run experiments on 5 datasets. The evaluator can test these datasets through the following steps:
\begin{itemize}
    \item Adult Census Income (ACI): Automatically downloaded via \texttt{shap.datasets.adult()} when running the code.
    \item Forest Cover Type (CovType): Automatically downloaded via \texttt{sklearn.datasets.fetch\_covtype()} when running the code.
    \item Bank Marketing (BM): Download from \href{https://www.kaggle.com/datasets/janiobachmann/bank-marketing-dataset}{Kaggle: Bank Marketing Dataset} and place \texttt{bank.csv} into \texttt{./data/}.
    \item Credit Card Fraud (CreditCard): Download from \href{https://www.kaggle.com/datasets/mlg-ulb/creditcardfraud/data}{Kaggle: Credit Card Fraud Detection} and place \texttt{creditcard.csv} in \texttt{./data/}.
    \item Download \texttt{HIGGS.csv.gz} from \href{https://archive.ics.uci.edu/ml/datasets/HIGGS}{UCI Machine Learning Repository: HIGGS} and extract \texttt{HIGGS.csv}. Then run the preprocessing script \texttt{data/HIGGS-preprocess.py} to generate \texttt{processed.pkl}. The preprocessing file is also available at \href{https://github.com/bartpleiter/tabular-backdoors}{GitHub: tabular-backdoors}. Finally, place \texttt{processed.pkl} in \texttt{./data/}.
    \item Download from \href{https://www.kaggle.com/datasets/hosseinah1/poker-game-dataset/data}{Kaggle: Poker Game Dataset} and place \texttt{poker-hand-training.csv} and \texttt{poker-hand-testing.csv} in \texttt{./data/}.
\end{itemize}


\subsection{Artifact Installation \& Configuration}
Our repo's README file contains installation instructions. The instructions are the following:

\begin{enumerate}
  \item \textbf{Clone the repository:}
  \begin{tcolorbox}[colback=black!5!white, colframe=black!75!white, boxrule=0.5pt]
  \footnotesize
  \begin{lstlisting}[language=bash]
git clone <repository-url>
cd catback
  \end{lstlisting}
  \end{tcolorbox}

  \item \textbf{Create a virtual environment, upgrade pip, and install the required dependencies:}
  \begin{tcolorbox}[colback=black!5!white, colframe=black!75!white, boxrule=0.5pt]
  \footnotesize
  \begin{lstlisting}[language=bash]
python -m venv env
. env/bin/activate
python -m pip install --upgrade pip
pip install -r requirements.txt
  \end{lstlisting}
  \end{tcolorbox}
\end{enumerate}

We have pinned the versions in requirements.txt to match the development environment for reproducibility.\footnote{By using the term reproducibility, we do not mean that the experiments will give identical results. We have not fixed any seed in our implementation for this matter. Moreover, it is worth noticing that even by fixing the seeds, the results will not be identical. This is due to the non-deterministic nature of operations in machine learning and also the hardware differences when running the experiments. Instead, by reproducible, we mean that the numbers will be in close range.} To resolve any issues with these specific versions (e.g., due to OS or Python version incompatibilities), the evaluator can install the latest stable versions by removing the \texttt{==version} part from the file or using \verb|pip install <package>| without versions.


\subsection{Major Claims}
Our major claim that can be reproduced by our public code is the following:
\begin{itemize}
    \item We applied our attack on five datasets and four models (both neural networks and classical machine learning methods), showing that it can generalize well in different settings. In particular, our attack reached $\approx$ 100\% attack success rate in most cases. This claim is supported by Table 1 in our paper.
\end{itemize}

\subsection{Evaluation}




To evaluate our experiments and reproduce Table~\ref{tab:exp-results} the evaluator can run our main script in the following way:

\begin{tcolorbox}[colback=black!5!white, colframe=black!75!white, boxrule=0.5pt]
\scriptsize
\begin{lstlisting}[language=bash]
python step_by_step.py --dataset_name <dataset> \
                       --model_name <model> \
                       --target_label <target_label> \
                       --epsilon <epsilon> 
\end{lstlisting}
\end{tcolorbox}

The possible values for the dataset argument are: ``aci'', ``bm'', ``higgs'', ``credit\_card'', ``covtype'', ``poker''. The values for the models are: ``ftt'', ``tabnet'', ``saint'', and ``xgboost''. Finally the values for epsilon are: 0.01, 0.02, 0.05, 0.1. Using all the possible combinations for these values will reproduce Table~\ref{tab:exp-results}.

The duration of running experiments highly depends on the combination of the chosen dataset, model, and hyperparameters. Also the environment and selected software and hardware (e.g., OS, GPU type) highly affects the duration. For example, with our settings, the experiments could finish in as little as 5 minutes when using the Bank Marketing dataset, whereas the HIGGS dataset had the longest running time and could take several hours to finish (e.g., with our settings, it took around 26 hours to finish one Tabnet experiment on HIGGS).

%
%
%
%




\end{document}